\documentclass[lettersize,journal]{IEEEtran}
\IEEEoverridecommandlockouts
\usepackage{amsmath,amsfonts}
\usepackage{array}
\usepackage[caption=false,font=normalsize,labelfont=sf,textfont=sf]{subfig}
\usepackage{textcomp}
\usepackage{stfloats}
\usepackage{xurl}
\usepackage{verbatim}
\usepackage{graphicx}
\usepackage{cite}

\usepackage{booktabs}
\usepackage{multirow}
\usepackage{wrapfig}
\usepackage{makecell}
\usepackage{caption}
\usepackage[caption=false]{subfig}
\usepackage{xspace}
\usepackage{enumitem}
\usepackage[most]{tcolorbox}
\usepackage[ruled,vlined,linesnumbered]{algorithm2e}

\newcommand{\modelname}{LLM-KGFR\xspace}

\begin{document}

\title{KGFR: A Foundation Retriever for Generalized Knowledge Graph Question Answering}


\markboth{Journal of \LaTeX\ Class Files,~Vol.~14, No.~8, August~2021}%
{Shell \MakeLowercase{\textit{et al.}}: A Sample Article Using IEEEtran.cls for IEEE Journals}



\author{Yuanning~Cui,
        Zequn~Sun, Wei~Hu,  and~Zhangjie~Fu*\thanks{* Corresponding author}
\IEEEcompsocitemizethanks{
    \IEEEcompsocthanksitem This work was supported by the National Natural Science Foundation of China (Nos. 62272219, 62406136, and U22B2062), the Jiangsu Provincial Science and Technology Major Project (No. BG2024042), the Natural Science Foundation of Jiangsu Province (No. BK20241246), and the Startup Foundation for Introducing Talent of NUIST (No. 1133142601017).
    \IEEEcompsocthanksitem Yuanning Cui is with the School of Computer Science, Nanjing University of Information Science and Technology, Nanjing 210044, China, and also with the State Key Laboratory for Novel Software Technology,
Nanjing University, Nanjing 210023, China
    (e-mail: yncui@nuist.edu.cn)
    \IEEEcompsocthanksitem Zequn Sun is with the State Key Laboratory for Novel Software Technology, Nanjing University, Nanjing 210023, China
    (e-mail: sunzq@nju.edu.cn)
    \IEEEcompsocthanksitem Wei Hu is with the State Key Laboratory for Novel Software Technology, Nanjing University, Nanjing 210023, China, and also with the National Institute of Healthcare Data Science, Nanjing University, Nanjing 210093, China
    (e-mail: whu@nju.edu.cn)
    \IEEEcompsocthanksitem  Zhangjie Fu is with the School of Cyber Science and Engineering, Nanjing University of Information Science and Technology, Nanjing 210044, China, and also with the Engineering Research Center of Digital Forensics, Ministry of Education, Nanjing University of Information Science and Technology, Nanjing 210044, China 
    (e-mail: fzj@nuist.edu.cn).
}
}

\maketitle

\begin{abstract}
Large language models (LLMs) excel at reasoning but struggle with knowledge-intensive questions due to limited context and parametric knowledge.
However, existing methods that rely on finetuned LLMs or GNN retrievers are limited by dataset-specific tuning and scalability on large or unseen graphs.
We propose the \textit{LLM–KGFR} collaborative framework, where an LLM works with a structured retriever, the \textit{Knowledge Graph Foundation Retriever} (\textit{KGFR}).
KGFR encodes relations using LLM-generated descriptions and initializes entities based on their roles in the question, enabling zero-shot generalization to unseen KGs.
To handle large graphs efficiently, it employs \textit{Asymmetric Progressive Propagation} (APP)—a stepwise expansion that selectively limits high-degree nodes while retaining informative paths.
Through node-, edge-, and path-level interfaces, the LLM iteratively requests candidate answers, supporting facts, and reasoning paths, forming a controllable reasoning loop.
Experiments demonstrate that \textit{LLM–KGFR} achieves strong performance while maintaining scalability and generalization, providing a practical solution for KG-augmented reasoning.
\end{abstract}



\section{Introduction}
\label{sec:introduction}

Large language models (LLMs) have shown impressive progress in natural language understanding and reasoning. Nevertheless, due to the finite scope of training corpora and the compression of knowledge into parameters, LLMs inevitably suffer from incomplete knowledge coverage and hallucinations~\cite{Hallucination}. To address this issue, external structured knowledge sources such as knowledge graphs (KGs)~\cite{TKDE_KBQA1, TKDE_KBQA2, TKDE_health_QA, TKDE_QA_survey1, TKDE_QA_survey2} offer an effective complement, providing factual grounding and enhancing reasoning reliability.

Despite their potential, integrating KGs into LLM-based question answering (QA) systems remains non-trivial.
First, KGs are large and heterogeneous, making it infeasible to directly feed their content into LLMs due to token and memory constraints.
Second, GNN-based approaches~\cite{G-Retriever, GNN-RAG} rely on KG-specific finetuning and full-graph message passing, which limits both generalization and scalability on large graphs.
Third, retrieval-augmented methods such as ToG~\cite{ToG} and RoG~\cite{RoG} depend on unstructured LLM queries or costly finetuning for relation-path generation, further constraining scalability and adaptability.
These limitations motivate the following research question:

\textit{How can we design a KG retrieval framework that generalizes to unseen graphs, scales to large KGs, and collaborates seamlessly with LLMs for reliable reasoning?}

To answer this question, we propose the \textbf{LLM–KGFR} collaborative framework, where \textbf{KGFR} (Knowledge Graph Foundation Retriever) serves as a question-conditioned graph retriever and works together with a frozen LLM. 
The overall framework is designed around three principles: generalization, scalability, and collaboration.

\textbf{First, generalization.}
KGs differ in domains, vocabularies, and relation schemas, posing a key challenge for cross-dataset generalization.
Many recent methods~\cite{G-Retriever, GNN-RAG} rely on dataset-specific LLM finetuning, where the model implicitly learns graph-specific patterns and thus struggles to transfer to new KGs without retraining.
In contrast, our KGFR adopts a \textit{question-conditioned initialization} that dynamically adapts entity embeddings according to the question context, assigning informative embeddings to mentioned entities and neutral ones to others.
Furthermore, the LLM generates unified textual relation descriptions that are encoded as structured representations, enabling meaningful embeddings even for unseen relations and supporting cross-KG generalization without any tuning.

\textbf{Second, scalability.}
Large KGs with millions of entities and edges make retrieval computationally intensive.
Traditional GNNs expand neighborhoods uniformly, causing combinatorial growth and high memory usage around hub nodes.
KGFR introduces an \textit{Asymmetric Progressive Propagation (APP)} mechanism that expands layer by layer from topic entities while selectively constraining high-degree nodes to limit redundant growth.
This asymmetric control retains informative links without inflating subgraphs, effectively balancing depth and breadth to achieve scalable retrieval on million-scale KGs.

\textbf{Third, collaboration.}
While KGFR ensures efficient retrieval, effective reasoning relies on dynamic interaction with the LLM.
To balance structural precision and semantic understanding, we design a \textit{controller–retriever loop}: the LLM acts as a controller that reflects on intermediate results, reformulates or raises follow-up queries when information is missing, and issues new retrieval requests, while KGFR executes them and returns structured evidence through node-, edge-, and path-level interfaces.
The LLM iteratively accesses candidate entities, supporting facts, and reasoning paths, while also generating unified relation descriptions that refine KGFR’s retrieval space.
This bidirectional collaboration enables controllable and interpretable reasoning beyond the capability of either component alone.

We evaluate our approach on seven QA benchmarks spanning diverse KG domains and reasoning tasks.
Experimental results show that our framework consistently achieves strong accuracy, generalizes robustly to unseen datasets, and scales effectively to million-entity graphs.

Our main contributions are summarized as follows:

\begin{itemize}
\item \textbf{Generalization:} We design KGFR with LLM-guided relation initialization and question-conditioned propagation, enabling it to handle unseen entities and relations across heterogeneous KGs without retraining.

\item \textbf{Scalability:} We propose an \textit{Asymmetric Progressive Propagation (APP)} mechanism that selectively constrains high-degree expansions, effectively controlling subgraph growth and ensuring scalable retrieval on large graphs.

\item \textbf{Collaboration:} We develop an \textit{LLM–KGFR} framework, where KGFR offers multi-level retrieval and the LLM conducts iterative reasoning through reflection and reformulation.
\end{itemize}

The remainder of this paper is organized as follows.
Section~\ref{sec:related_work} reviews related work.
Section~\ref{sec:kgfr} introduces the Knowledge Graph Foundation Retriever (KGFR), and Section~\ref{sec:collaborative_question_answering} presents the LLM–KGFR collaborative framework.
Section~\ref{sec:experiments} reports experiments, and Section~\ref{sec:conclusions} concludes the paper.

\section{Related Work}
\label{sec:related_work}

\paragraph{LLM reasoning with KGs}
LLMs often hallucinate on knowledge-intensive tasks~\cite{Hallucination}, prompting the use of structured KGs for factual grounding and improved reliability.
Existing studies fall into two categories: (1) constructing task-specific graphs from texts to enhance retrieval and summarization~\cite{GraphRAG, LightRAG, HippoRAG, GFM-RAG}, and (2) leveraging open KGs as external evidence for natural-language questions.
Within the first line, GFM-RAG~\cite{GFM-RAG} employs a graph foundation model to improve text retrieval and summarization, but its task objective, input–output form, and benchmarks differ substantially from KG-based QA, making it not directly comparable to our work.
Representative methods in the second line include KD-CoT~\cite{KD-CoT}, which retrieves KG facts to guide chain-of-thought reasoning, and agent-style frameworks such as StructGPT~\cite{StructGPT} and ToG~\cite{ToG}, where LLMs interact with KGs to explore reasoning paths.
EffiQA~\cite{EffiQA} further introduces a compact plug-in retriever to efficiently explore entities and relations.
LightPROF~\cite{LightPROF} encodes KG structures into soft prompts via a lightweight adapter.
RoG~\cite{RoG} and GCR~\cite{GCR} adopt planning–retrieval–reasoning pipelines that return KG paths as structured evidence.
To better exploit graph topology, GNN-RAG~\cite{GNN-RAG} integrates GNN-based retrieval, and G-Retriever~\cite{G-Retriever} applies GNN-based prompt tuning; however, both rely on KG-specific training or finetuned LLMs, which hinders cross-KG generalization and scalability.

\paragraph{KG foundation models}
Early graph foundation models (GFMs) follow a pretraining–finetuning paradigm to transfer across datasets~\cite{GCL2, MCM1, MCM2, GCL1, GraphMAE}. 
Prompt-based GFMs~\cite{GPPT, AllInOne, GraphPrompt, PGCL, SGL-PT, Deep-GPT, HetGPT, GraphPromptReview} further improve adaptability via universal prompt templates for diverse graph-level, edge-level, and node-level tasks. 
Recently, several KG-oriented foundation models have targeted relational reasoning and KG completion~\cite{PR4LP, ULTRA, KG-ICL, TRIX}. 
These approaches primarily focus on graph analytics or completion objectives, whereas our setting centers on natural-language QA over KGs with a collaborative LLM–retriever workflow.

\section{Preliminaries}
\label{sec:pre}

\paragraph{Knowledge graph question answering} KGQA is the task of answering natural language questions based on the facts in a given KG.
The background KG is formulated as $\mathcal{G}=(\mathcal{E}, \mathcal{R}, \mathcal{T})$, where $\mathcal{E}$, $\mathcal{R}$, and $\mathcal{T}$ are the sets of entities, relations, and facts, respectively.
A fact, formed as $(s, r, o)\in\mathcal{T}$, represents a directed edge between entities, where $s$ and $o$ are the subject and object entities from $\mathcal{E}$, respectively, and $r$ is the relation between them.
A question is formed as $(q, \mathcal{A}_q, \mathcal{P}_q)$, where $q$ is the natural language question, $\mathcal{A}_q\subseteq\mathcal{E}$ is the set of answer entities, and $\mathcal{P}_q\subseteq\mathcal{E}$ is the set of topic entities.
Each question contains at least one topic entity, which anchors the natural language question to the KG. 
The KG facts about the topic entities in the question are the basis for answering the question.
For complex questions, KGQA methods should be capable of reasoning over the subgraphs surrounding the topic entities, filtering out irrelevant information, and aggregating useful relational facts.

\paragraph{Generalized KGQA} 
Generalized KGQA aims to answer questions over both seen and unseen KGs.
In the generalized KGQA task, during the reasoning phase, a KG $\mathcal{G}' = (\mathcal{E}', \mathcal{R}', \mathcal{T}')$ is given, which may differ from the one used in the training set. 
A generalized KGQA question is formalized as $(q, \mathcal{A}_{q}, \mathcal{P}_{q}, \mathcal{H}_{q})$, where $q$ and $\mathcal{P}_q\subseteq\mathcal{E}'$ are defined in the same way as in KGQA. 
$\mathcal{H}_q$ represents a set of candidate answers, and $\mathcal{A}_q\subseteq\mathcal{H}_q$ is the set of correct answers. 
For example, in multiple-choice questions, $\mathcal{H}_q$ is the set of all options, while $\mathcal{A}_q$ contains the correct options. 
KGQA can be seen as a special case of generalized KGQA, where the candidate answers are all entities, i.e., $\mathcal{H}_q = \mathcal{E}$.
The original KG is a directed graph.
To enhance its connectivity, following the convention \cite{TransE,  EmbedKGQA}, we incorporate inverse relations and facts into the KG.
Specifically, for every fact $(s, r, o)\in\mathcal{T}$, we introduce a reverse fact $(o, r^{-1}, s)$, where $r^{-1}$ is the inverse relation of $r$.
In the following sections, we assume that inverse relations and facts are included in $\mathcal{R}$ and $\mathcal{T}$.

\section{Knowledge Graph Foundation Retriever}
\label{sec:kgfr}
\begin{figure*}
  \centering
  \includegraphics[width=0.95\linewidth]{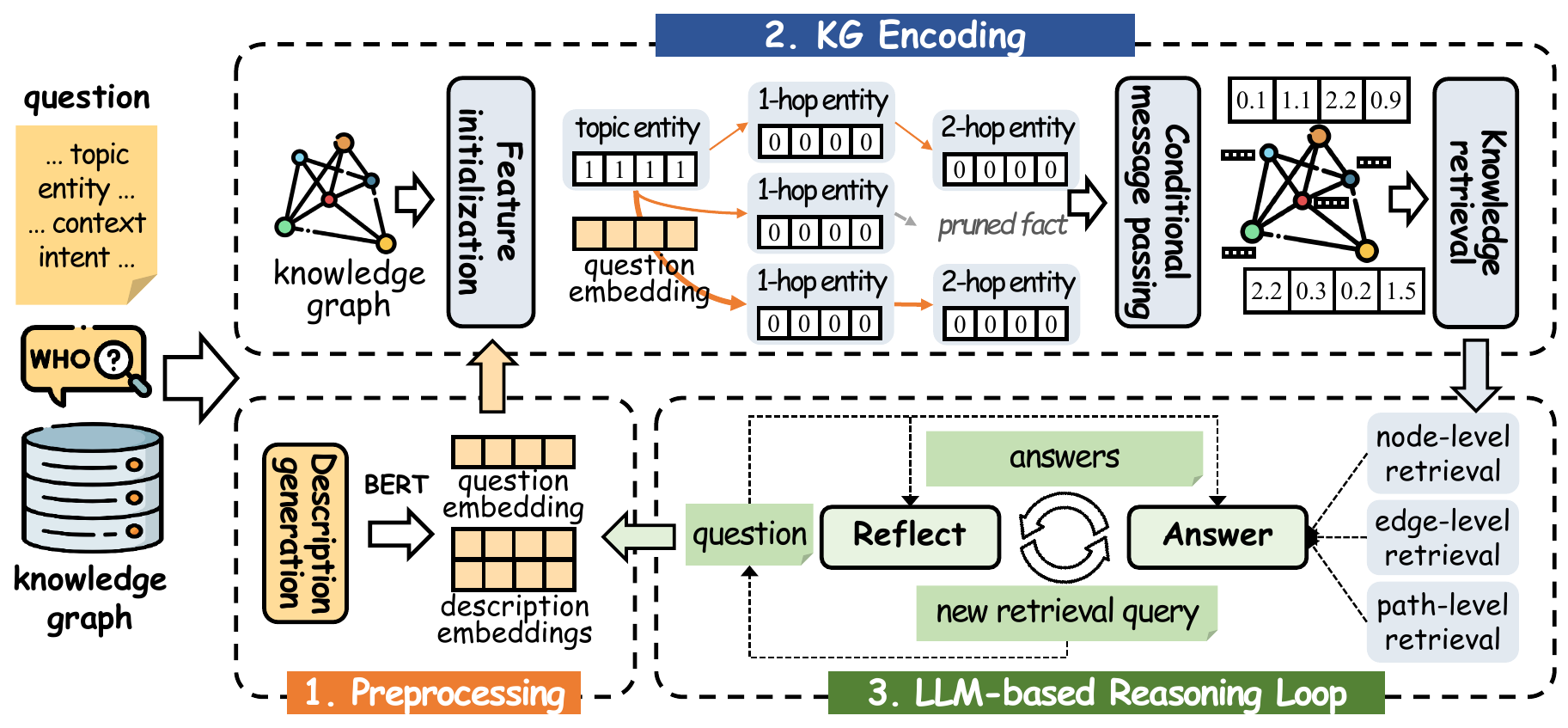}
  \caption{Framework of LLM-KGFR.
Given a KG, we first perform preprocessing by generating relation descriptions and encoding them using BERT.
Given a question, the model performs asymmetric progressive propagation from the topic entity.
It then conducts multi-level (node, edge, path) retrieval to iteratively generate and refine answers with the LLM, terminating when the answer is confirmed or the maximum number of steps is reached.}
  \label{fig:details}
\end{figure*}

Figure~\ref{fig:details} presents an overview of the framework.
We propose the Knowledge Graph Foundation Retriever (KGFR), a question-conditioned retriever for generalized and scalable knowledge retrieval without finetuning.
For \textbf{generalization}, KGFR employs language-guided initialization, encoding relations from LLM-generated descriptions and initializing entities by their roles in the question.
For \textbf{scalability}, it performs asymmetric progressive propagation from topic entities while constraining high-degree nodes.
For \textbf{collaboration}, it exposes node-, edge-, and path-level interfaces that interact with the LLM for iterative reasoning and reflection.

\subsection{Relation Initialization via Unified Descriptions}
\label{sec:relation_init}

A key challenge to \textit{generalization} in KGQA is the inconsistent naming and formatting of relations across KGs.
Semantically equivalent relations often appear under different identifiers or conventions (e.g., Freebase \texttt{/location/country/capital}, Wikidata \texttt{P36}, or natural-language forms such as ``is the capital of''), making relation names unreliable indicators of semantics.
As a result, retrievers trained on one KG struggle to transfer to others with distinct relation vocabularies.

To address this, KGFR leverages LLMs to generate \textit{unified textual descriptions} for relations from their names and a few example triples.
These descriptions abstract away dataset-specific identifiers and capture consistent relation semantics in natural language, which are then encoded as initial relation embeddings.
An illustrative prompt is shown in Figure~\ref{fig:prompt}. 

\begin{figure}[!ht]
\centering
\begin{tcolorbox}[colback=gray!10,colframe=gray!70,title=Prompt for Relation Description Generation,width=0.98\linewidth]
\textbf{Task:} Generate a description of the given relation.\\[2pt]
\textbf{Relation:} \texttt{sports.sport.teams}\\[2pt]
\textbf{Examples:} (Basketball, sports.sport.teams, Los Angeles Lakers); $\cdots$\\[2pt]
\textbf{Output Example:} \texttt{sports.sport.teams} describes how a sport is associated with the teams that participate in it.
\end{tcolorbox}
\caption{Illustrative prompt used for generating unified textual descriptions of relations.}
\label{fig:prompt}
\end{figure}

By normalizing relations into such unified descriptions, KGFR aligns heterogeneous schemas with natural language questions, enabling question-conditioned message passing and robust cross-KG generalization.

\subsection{Question-Conditioned Message-Passing}
\label{sec:conditional_message_passing}

Our KGFR's message-passing mechanism addresses two key challenges: enabling natural language understanding and ensuring strong generalization to unseen graphs.
Specifically, we integrate the BERT-encoded question representations into both relation embedding initialization and attention computation, enabling the KGFR to process linguistic inputs. 
To ensure generalization to unseen KGs, we employ non-learnable entity vectors \cite{ULTRA, KG-ICL, TRIX, NBFNet, REDGNN} during this process.

\textbf{Feature initialization.}
Given a question $q$, a KG $\mathcal{G}=\{\mathcal{E}, \mathcal{R}, \mathcal{T}\}$, and a topic entity set $\mathcal{P}_q$, we first initialize the features of relations and entities. 

For relations, we initialize their features using the textual embeddings of their descriptions:
\begin{equation}
    \mathbf{r}^{(0)} = \operatorname{BERT}(u_{r}),
    \label{eq:relation_embedding_initialization}
\end{equation}
where $\operatorname{BERT}()$ denotes the BERT encoder~\cite{BERT}, and $u_r$ is the textual description of the relation $r$.

For entities, KGFR adopts a lightweight yet robust initialization scheme widely used in recent KG foundation models~\cite{ULTRA, KG-ICL, TRIX}.  
Instead of encoding millions of entity names—which is computationally expensive and often inconsistent across KGs—topic entities are assigned fixed one-vectors $\mathbf{1}_d$, and all others zero-vectors $\mathbf{0}_d$, where $d$ is the embedding dimension.  
This initialization avoids dependency on noisy textual features, maintains efficiency on large-scale KGs, and has been empirically shown to support generalization to unseen graphs.

\textbf{Question-conditioned message-passing.}
Next, we propagate messages through the graph to update the embeddings of the relations and entities. 
In the $i$-th layer, we first update the embeddings of the relations as follows:
\begin{equation}
    \mathbf{r}^{(i+1)} = \mathbf{W}_1^{(i)}\left[\mathbf{r}^{(i)}\,;\,\operatorname{BERT}(q)\right],
\end{equation}
where $\mathbf{W}_1^{(i)}\in\mathbb{R}^{d\times 2d}$ is a learnable weight matrix, and $[\,;\,]$ is the concatenation operation.
In this way, we retain the message from the previous layer while injecting question-conditioned information at each layer.
Then, we adopt a progressive propagation \cite{REDGNN} to update the embeddings of the entities as follows:
\begin{align}
    \mathbf{e}^{(i+1)} = \sum_{(s, r, e)\in \mathcal{N}^{(i)}_{\mathcal{P}_q}(e)}\mathbf{W}_2^{(i)}\operatorname{MSG}^{(i)}(s, r, q),
\label{eq:update_entity}
\end{align}
where $\mathbf{W}_2^{(i)}\in\mathbb{R}^{d\times d}$ is a learnable weight matrix, and $\mathcal{N}^{(i)}_{\mathcal{P}_q}(e)$ is the set of facts that both includes entity $e$ and is within the $i$-hop neighbors of the topic entities.
$\operatorname{MSG}^{(i)}()$ is the message function:
\begin{align}
    \operatorname{MSG}^{(i)}(s, r, q) = \alpha_{s;r;q}^{(i)}\left(\mathbf{s}^{(i)}+\mathbf{r}^{(i)}\right),
\end{align}
where $\mathbf{s}^{(i)}$ and $\mathbf{r}^{(i)}$ are the $i$-th layer embeddings of $s$ and $r$, respectively.
$\alpha_{s;r;q}$ is the attention of the edge, which is calculated by
\begin{equation}
\alpha_{s;r;q}^{(i)} = f\left(\mathbf{W}_3^{(i)}g\left(\mathbf{W}_4^{(i)}\mathbf{s}^{(i)}+\mathbf{W}_5^{(i)}\mathbf{r}^{(i)}+\mathbf{W}_6^{(i)}\mathbf{q}\right)\right),
\label{eq:attention}
\end{equation}
where $f()$ and $g()$ denote the activation functions $\operatorname{sigmoid}$ and $\operatorname{ReLU}$, respectively.
$\mathbf{W}_3^{(i)}\in\mathbb{R}^{1\times d_\text{attn}}$, $\mathbf{W}_4^{(i)}$, $\mathbf{W}_5^{(i)}$, and $\mathbf{W}_6^{(i)}\in\mathbb{R}^{d_\text{attn}\times d}$ are learnable weight matrices, where $d_{\text{attn}}$ is a hyperparameter that reduces dimensionality.
$\mathbf{q}=\operatorname{BERT}(q)$. 

\subsection{Asymmetric Progressive Propagation}
\label{sec:APP}

To ensure scalability on large KGs, we design the \textit{Asymmetric Progressive Propagation} (APP) mechanism, which integrates two complementary principles: \textbf{progressive expansion} and \textbf{asymmetric pruning}.

\textbf{Progressive expansion.}
Natural-language questions often imply a multi-hop reasoning process.  
APP follows this intuition by starting from the topic entities $\mathcal{P}_q$ and expanding one hop at a time.  
At each hop, the newly reached edges are merged with previously retrieved ones to form a progressively enlarged retrieval subgraph.  
This progressive expansion keeps the retrieval scope localized around relevant entities instead of the entire KG, enabling efficient large-scale deployment.

\textbf{Asymmetric pruning.}
Naive propagation tends to suffer from high-degree relations that cause uncontrolled growth.  
For instance, entities such as \textit{China} may participate in relations like \textit{(China, citizens, ?)} that connect to millions of nodes, most of which are irrelevant.  
However, pruning the entity itself would also remove useful edges such as \textit{(China, capital, ?)} or \textit{(China, official language, ?)}.  
APP therefore performs pruning at the \((s, r, ?)\) level: when a relation type yields excessive neighbors, further expansion along that relation is suppressed, while other relations of $s$ remain available for propagation.  
This asymmetric rule effectively controls hub-induced explosion without discarding informative reasoning paths.

\textbf{Formalization.}  
Let $\mathcal{N}^{(i)}_{q}(e)$ denote the set of edges reachable from entity $e$ at hop $i$.  
Progressive propagation is defined as
\begin{equation}
\mathcal{N}^{(i+1)}_{q}(e) \;=\; 
\mathcal{N}^{(i)}_{q}(e)\; \cup 
\bigcup_{(e,r,o)\in \mathcal{T}} \mathrm{Expand}(e,r),
\label{eq:propagation}
\end{equation}
where $\mathrm{Expand}(e,r)$ inserts edges $(e,r,o)$ into the retrieval frontier.  
Define $C_{e,r}=\{\,o\mid(e,r,o)\in\mathcal{T}\,\}$ as the candidate neighbor set. Then
\begin{equation}
\mathrm{Expand}(e,r) =
\begin{cases}
\{(e,r,o)\mid o\in C_{e,r}\}, & |C_{e,r}| \leq \lambda, \\[4pt]
\{(e,r,o)\mid o \in \mathcal{S}_i\}, & |C_{e,r}| > \lambda,
\end{cases}
\label{eq:expand}
\end{equation}
where $\mathcal{S}_i
= \bigcup_{i=0}^i \Big\{\, x \;\Big|\; \exists (u,r,v)\in \mathcal{N}^{(i)}_{q}(u),\; x\in\{u,v\} \,\Big\}
$  
is the cumulative set of entities reached up to hop $i$. 
$\lambda$ is a threshold controlling the maximum number of neighbors expanded per relation, preventing high-degree nodes from overwhelming the propagation.

\subsection{Pre-training Objective}
\label{sec:pre-training}

Since we have already incorporated the question as a condition in the question-conditioned message-passing process, we directly read the entity embeddings from the last layer to compute their scores:
\begin{equation}
    c_{e \mid q} = \mathbf{W}_7\mathbf{e}^{(L)},
    \label{eq:score_function}
\end{equation}
where $\mathbf{W}_7\in\mathbb{R}^{1\times d}$ is a learnable weight matrix, and $L$ is the number of message-passing layers.

Given the background KG $\mathcal{G}=(\mathcal{E}, \mathcal{R}, \mathcal{T})$ and a set of question-answering training data $\mathcal{D}=\{(q_1, \mathcal{A}_{q_1}, \mathcal{P}_{q_1}), \dots,$ $ (q_n, \mathcal{A}_{q_n}, \mathcal{P}_{q_n})\}$, we introduce a variant of the multi-class log-loss function \cite{multi_class_log_loss} to pre-train the KGFR:
\begin{equation}
\resizebox{0.9\linewidth}{!}{$
\begin{aligned}
\mathcal{L}
= \sum_{(q_i,\mathcal{A}_{q_i},\mathcal{P}_{q_i})\in\mathcal{D}}
\Big[
  \underbrace{\log\!\sum_{x\in\mathcal{E}} \exp(c_{x \mid q_i})}_{\text{All candidates (denominator)}}
  \;-\;
  \underbrace{\log\!\sum_{a\in\mathcal{A}_{q_i}} \exp(c_{a \mid q_i})}_{\text{Positive set (numerator)}}
\Big].
\end{aligned}
$}
\end{equation}

\section{Collaborative Question Answering with LLMs}
\label{sec:collaborative_question_answering}

While the previous section establishes KGFR as a general and scalable retriever, effective question answering requires it to operate in concert with a language model capable of reasoning over retrieved evidence.
This section introduces the \textbf{LLM–KGFR collaborative framework}, where the LLM and KGFR jointly perform QA through a controller–executor interaction loop (Figure~\ref{fig:details}).
The LLM interprets questions, formulates relation descriptions, and decides when additional retrieval is needed, while KGFR executes APP-based propagation and returns structured evidence at node, edge, and path levels.
Through iterative retrieval, reasoning, and reflection, this collaboration enables large language models to reason over vast KGs efficiently and transparently.

\subsection{KGFR-based Knowledge Retrieval}
\label{sec:KGFR}

To enable collaborative QA, KGFR exposes structured retrieval interfaces that let the LLM acquire information at different granularities.  
Rather than returning raw neighbors or invoking additional models, KGFR organizes its outputs into three complementary levels: \textbf{node-level} candidate entities, \textbf{edge-level} supporting facts, and \textbf{path-level} connections to topic entities.  
These retrievals operate on the \textit{retrieval subgraph} $\mathcal{G}^{(L)}_q=(\mathcal{S}_L,\mathcal{N}^{(L)}_q)$ produced by APP after $L$ hops, where $\mathcal{S}_L$ is the set of reached entities and $\mathcal{N}^{(L)}_q$ the set of reached edges.  
This design allows the LLM to flexibly request coarse-to-fine evidence depending on the reasoning stage, while preserving scalability and generalization.

\textbf{Node-level retrieval.}  
KGFR ranks entities within the current subgraph and returns the top-$k$ candidates:
\begin{equation}
\mathcal{C}_q = \operatorname{Top}_k\{\, e : c_{e \mid q} \;\mid\; e\in \mathcal{S}_L \,\},
\label{eq:kgfr_node}
\end{equation}
where $c_{e \mid q}$ is the question-conditioned score of entity $e$.  
This narrows the search space of a large KG into a concise, ranked set of plausible answers for the LLM to examine.

\textbf{Edge-level retrieval.}  
To provide factual grounding, KGFR retrieves the most relevant edges for an entity $e$ within the subgraph:
\begin{equation}
\mathcal{I}_{q;e} = \operatorname{Top}_n\{\, (s,r,e) : \alpha_{s;r;q}^{\max} \;\mid\; (s,r,e)\in \mathcal{N}^{(L)}_q \,\},
\label{eq:kgfr_edge}
\end{equation}
where $\alpha_{s;r;q}^{\max} = \max_{1\le i\le L}\alpha^{(i)}_{s;r;q}$ is the maximum attention weight across message-passing hops.  
Let $\mathcal{I}_q = \bigcup_{e\in \mathcal{C}_q}\mathcal{I}_{q;e}$ be the set of relevant edges.

\textbf{Path-level retrieval.}  
To reveal multi-hop reasoning chains, KGFR computes the shortest paths in $\mathcal{G}^{(L)}_q$ from each important entity to every topic entity in $\mathcal{P}_q = \{e_q^{(1)}, e_q^{(2)}, \dots\}$:
\begin{equation}
\mathcal{P}^{\text{path}}_q =
\bigcup_{e \in \mathcal{C}_q}
\bigcup_{e_q \in \mathcal{P}_q}
\operatorname{SP}(e, e_q;\,\mathcal{G}^{(L)}_q),
\label{eq:kgfr_path}
\end{equation}
where $\operatorname{SP}$ returns directed shortest paths between $e$ and $e_q$ within the subgraph $\mathcal{G}^{(L)}_q$.  
This deterministic retrieval avoids training path-generating LLMs while ensuring efficiency, interpretability, and transferability to unseen KGs.

Together, node-, edge-, and path-level retrieval grant the LLM modular access to KG evidence—candidates to narrow the space, supporting facts to ground decisions, and paths to expose reasoning chains—enabling scalable and robust collaborative QA.

\subsection{LLM-based Generation and Reflection}
\label{sec:LLM}

The LLM in the LLM–KGFR framework plays a controlling role in driving the QA process, complementing KGFR’s structured retrieval with language-based reasoning.  
Its role spans three dimensions: generating and aligning relation representations, reflecting on intermediate answers, and adaptively refining retrieval.

\textbf{Generation and fact verbalization.}
Since language and graph representations differ substantially, effective communication requires a unified representational space bridging the two modalities.
First, the LLM generates \emph{relation descriptions} for each relation (Section~\ref{sec:relation_init}); these unified textual descriptions serve as anchors for initializing relation embeddings in KGFR.
Separately, the LLM also induces \emph{verbalization templates} for each relation. Subsequently, when KGFR retrieves factual triples $(s, r, o)$, these templates are applied to verbalize the structured facts into natural sentences, allowing the evidence to be seamlessly integrated into subsequent reasoning steps.
Together, this two-part design—unified relation descriptions for initialization and relation-specific verbalization templates for natural-language rendering—enables KGFR to align with linguistic semantics and helps the LLM accurately interpret graph-based retrieval results.

\textbf{Answer generation and reflection.}
Given the candidate entities, supporting facts, and reasoning paths retrieved by KGFR, the LLM synthesizes a coherent natural-language answer through contextual reasoning and factual aggregation.
After each generation round, it performs a \emph{reflection step} to evaluate the sufficiency and consistency of the produced answer.
If the retrieved evidence adequately supports the conclusion and no contradictions are detected, the reasoning cycle terminates.
Otherwise, the LLM analyzes which parts of the reasoning chain remain uncertain or underspecified, identifies missing entities or relations, and formulates a targeted follow-up query to KGFR.

\textbf{Adaptive question rewriting and entity focusing.}  
When retrieval is incomplete, the LLM can rewrite the original question into auxiliary sub-questions $\mathcal{Q}_q$ to guide further retrieval~\cite{ToG,rewrite1,rewrite2}.  
It also identifies key entities from the current reasoning context and directs KGFR to concentrate subsequent retrieval on these entities.  
The rewritten questions and selected entities are then fed into the next iteration, enabling progressively refined reasoning.

\subsection{Reasoning Pipeline}
\label{app:pipeline}

For clarity, we describe the complete reasoning workflow of the LLM–KGFR collaboration below.
Algorithm~\ref{alg:pipeline} outlines the full reasoning pipeline of the framework, which proceeds in two stages given a question $q$, topic entities $\mathcal{P}_q$, and a KG $\mathcal{G}=(\mathcal{E},\mathcal{R},\mathcal{T})$.

\emph{Initial retrieval.}  
The LLM first generates unified textual descriptions for all relations, which are encoded by a frozen BERT encoder to initialize relation embeddings (Section~\ref{sec:relation_init}).  
Entities are initialized using KGFR’s initialization strategy described in Section~\ref{sec:conditional_message_passing}. 
KGFR then performs question-conditioned propagation with APP and produces three types of evidence within the retrieval subgraph:  
(i) node-level candidate entities $\mathcal{C}_q$,  
(ii) edge-level important facts $\mathcal{I}_q$, and  
(iii) path-level connections $\mathcal{P}^{\text{path}}_q$ obtained by shortest paths from retrieved entities to topic entities.  
This stage builds a compact, question-specific subgraph without any dataset-specific training, ensuring both efficiency and generalization.

\emph{Iterative retrieval.}  
Based on the retrieved evidence, the LLM synthesizes answers and performs reflection.  
If the evidence is insufficient, the LLM may (i) rewrite the question into sub-questions $\mathcal{Q}_q$ to trigger further retrieval, or (ii) identify key entities to focus additional edge- or path-level exploration.  
This reasoning loop continues until the LLM indicates that a final answer has been reached based on sufficient evidence, or a predefined maximum number of steps is reached.

\begin{table*}
\centering
    \caption{Dataset statistics. ``S'' and ``C'' denote ``Support'' and ``Counter'', respectively.}
    \vskip 6pt
    \label{tab:dataset_KBQA}
    \begin{tabular}{c|l|rrrrrr|l}
    \toprule
    Answer types & Datasets & Entity & Relation & Fact & Training & Development & Testing & KGs \\
    \midrule
    \multirow{3}{*}{Entities}
         & WebQSP & 1,298,306 & 6,094 & 3,791,303 & 2,848 & 250 & 1,639 & Freebase \\
         & CWQ & 2,259,510 & 6,649 & 7,269,449 & 27,639 & 3,519 & 3,531 & Freebase \\
         & GrailQA & 430,781 & 6,484 & 2,408,034 & - & - & 1,000 & Freebase \\
    \midrule
    S / C & ExplaGraphs & 6,331 & 28 & 9,771 & - & - & 2,766 & ConceptNet \\
    \midrule
    \multirow{3}{*}{Choices}
    & CSQA & 34,869 & 16 & 257,767 & - & - & 1,241 & ConceptNet\\
    & OBQA & 22,961 & 16 & 173,102 & - & - & 500 & ConceptNet\\
    & MedQA & 3,364 & 15 & 15,265 & - & - & 1,273 & UMLS \& DrugBank\\
    \bottomrule
    \end{tabular}
\end{table*}

\begin{algorithm}
\caption{Reasoning Pipeline of LLM–KGFR}
\label{alg:pipeline}
\KwIn{Question $q$, topic entities $\mathcal{P}_q$, and KG $\mathcal{G}=(\mathcal{E},\mathcal{R},\mathcal{T})$.}
\KwOut{Predicted answers $\mathcal{A}_{q;\text{pred}}$.}

\tcc{Stage 1: Initial retrieval}
Encode descriptions with BERT to initialize relation embeddings\;
Initialize entity embeddings\;
Run question-conditioned APP propagation\;
Retrieve candidate entities $\mathcal{C}_q$\;
Retrieve important facts $\mathcal{I}_q$ around $\mathcal{P}_q\cup\mathcal{C}_q$\;
Compute path-level connections $\mathcal{P}^{\text{path}}_q$ via shortest paths\;

\tcc{Stage 2: Iterative retrieval}
\For{$step \leftarrow 1$ \KwTo $max\_steps$}{
  Generate answer candidates $\mathcal{A}_{q;\text{pred}}$ (LLM)\;
  Reflect on evidence sufficiency and answer consistency\;
  \If{evidence insufficient}{
    \If{LLM produces sub-questions $\mathcal{Q}_q$}{
        Trigger new retrieval for $\mathcal{Q}_q$\;
    }
    \If{LLM identifies key entities}{
        Focus additional edge retrieval around selected entities\;
    }
  }
  \If{LLM signals sufficient evidence}{
    \textbf{break}\;
  }
}
\Return final answers $\mathcal{A}_{q;\text{pred}}$.
\end{algorithm}

\section{Experiments and Results}
\label{sec:experiments}

Our evaluation is guided by four research questions (RQs):
\begin{itemize}
    \item \textit{RQ1 (Performance):} How effectively does \modelname perform on KGQA compared with existing methods?
    \item \textit{RQ2 (Generalization):} Can \modelname generalize to unseen KGs and QA datasets with heterogeneous schemas?
    \item \textit{RQ3 (Scalability and Efficiency):} Does \modelname enable efficient inference on large KGs while preserving accuracy?
    \item \textit{RQ4 (Module Effectiveness):} What is the contribution of each proposed module to the overall performance?
\end{itemize}

We evaluate \modelname on seven KGQA benchmarks covering diverse domains, scales, and question styles. The source code is available at \url{https://github.com/yncui-nju/KGFR}.

\subsection{Settings}
\label{sec:settings}

\paragraph{Datasets}
We evaluate \modelname\ on seven datasets that span factual, compositional, and commonsense reasoning, as summarized in Table~\ref{tab:dataset_KBQA}.

WebQSP \cite{WebQSP} contains 4,737 natural language questions requiring up to 2-hop reasoning, 
while CWQ \cite{CWQ} includes 34,699 complex questions involving up to 4 hops. 
Both are built upon Freebase as the background KG. 
Following prior work \cite{RoG, UniKGQA, PullNet, NSM, GraftNet, GNN-RAG}, 
we first extract local subgraphs centered on topic entities for each dataset. 
These subgraphs are then merged into dataset-specific background KGs used during inference, 
whose overall scales remain comparable to those adopted in previous large-KG QA studies \cite{largeEA, AceKG, CS-KG}.

GrailQA \cite{GrailQA} further evaluates generalization to unseen domains and compositional query structures. 
It is also based on Freebase and undergoes the same subgraph extraction and merging procedure. 
Following \cite{PoG}, we do not perform additional training on GrailQA and directly evaluate 
the retriever pre-trained on CWQ in a zero-shot setting.

ExplaGraphs \cite{ExplaGraphs, G-Retriever} provides explanation graphs for determining whether two arguments are supportive or contradictory.
CommonSenseQA (CSQA) \cite{CSQA} and OpenBookQA (OBQA) \cite{OBQA} are multiple-choice QA datasets built on ConceptNet \cite{ConceptNet}, with 1,241 and 500 test questions (five and four options, respectively) following \cite{KagQA, QA-GNN}.
MedQA \cite{MedQA} is a biomedical multiple-choice dataset (four options) whose background KG integrates UMLS \cite{UMLS} and DrugBank \cite{DrugBank}.

\paragraph{Implementation details}
KGFR adopts a 4-layer message-passing backbone.
The threshold of the asymmetric pruning module is set to $\lambda{=}100$.
We use BGE-Large-EN-v1.5\footnote{\url{https://huggingface.co/BAAI/bge-large-en-v1.5}} as the BERT encoder to encode both the question and the relation descriptions, keeping encoder parameters frozen during all experiments.
The hidden dimension and attention head count are set to $d{=}1024$ and $d_{\text{attn}}{=}4$, respectively.
Both the top-$k$ candidate size and neighbor selection $n$ are set to 20.
The LLM reasoning loop is limited to three iterations.
Optimization uses the Adam optimizer with a learning rate of $1\mathrm{e}{-4}$, a maximum of 200 epochs, and early stopping with patience 5.
In the main evaluation, we adopt four LLMs—Qwen-max,\footnote{\url{https://qwen.ai/blog?id=qwen2.5-max}} GPT-4o-mini,\footnote{\url{https://platform.openai.com/docs/models/gpt-4o-mini}} GPT-4-Turbo,\footnote{\url{https://platform.openai.com/docs/models/gpt-4-turbo}} and GPT-4\footnote{\url{https://platform.openai.com/docs/models/gpt-4}}—and report results on WebQSP and CWQ.
For cost considerations, subsequent analyses adopt Qwen-max and GPT-4o-mini.
For ExplaGraphs, CSQA, OBQA, and MedQA, we use the frozen KGFR pre-trained on CWQ.
Model pre-training was done on a workstation featuring two Intel Xeon Gold CPUs, four NVIDIA A800 (80 GB) GPUs, and Ubuntu 18.04 LTS, whereas all evaluation was carried out on a server with four NVIDIA A6000 (48 GB) GPUs.
The retriever has 28M parameters.
Relation descriptions are generated once during data preprocessing and cached for reuse. 
We adopt GPT-4o-mini as the description generator.
For each relation, we construct a prompt by providing five example triples involving that relation as context, 
which helps the LLM produce more informative and grounded descriptions. 
The generated descriptions are stored and reused across all queries, 
avoiding redundant computation.

\begin{table*}[!bht]
\setlength{\tabcolsep}{8pt}
\centering
\caption{Results on WebQSP and CWQ. The best score on each metric is in \textbf{bold}, and the second best score is \underline{underlined}.}
\vskip 6pt
\label{tab:main_results}
{\small
\begin{tabular}{ll|ccc|ccc}
\toprule
\multirow{2}{*}{Types} & \multirow{2}{*}{Methods}
& \multicolumn{3}{c|}{WebQSP} & \multicolumn{3}{c}{CWQ} \\ 
\cmidrule(lr){3-5} \cmidrule(lr){6-8}
& & F1 & Hit & H@1 & F1 & Hit & H@1 \\ 
\midrule
\multirow{4}{*}{Embedding}
& KV-Mem \cite{KV-Mem} & 38.6 & 46.7 & - & - & 21.1 & - \\
& EmbedKGQA \cite{EmbedKGQA} & - & 66.6 & - & - & 45.9 & - \\
& TransferNet \cite{TransferNet} & - & 71.4 & - & - & 48.6 & - \\
& Rigel \cite{Rigel} & - & 73.3 & - & - & 48.7 & - \\
\midrule
\multirow{8}{*}{GNN}
& GraftNet \cite{GraftNet} & 60.4 & 66.4 & - & 32.7 & 36.8 & - \\
& PullNet \cite{PullNet} & - & 68.1 & - & - & 45.9 & - \\
& NSM \cite{NSM}& 62.8 & 68.7 & - & 42.4 & 47.6 & - \\
& SR\,+\,NSM (+\,E2E) \cite{SR+NSM} & 64.1 & 69.5 & - & 47.1 & 50.2 & - \\
& NSM\,+\,h \cite{NSM+h} & 67.4 & 74.3 & - & 44.0 & 48.8 & - \\
& SQALER \cite{SQALER} & - & 76.1 & - & - & - & - \\
& UniKGQA \cite{UniKGQA} & 72.2 & 77.2 & - & 49.1 & 51.2 & - \\
&ReaRev\,+\,LMsr \cite{ReaRev} & 72.8 & 77.5 & - & 49.7 & 53.3 & - \\
\midrule
\multirow{5}{*}{KG\,+\,LLM}
& KD-CoT  \cite{KD-CoT}& 52.5 & 68.6 & - & - & 55.7 & - \\
& StructGPT  \cite{StructGPT} & - & 72.6 & - & - & 54.3 & - \\
& KB-BINDER \cite{KB-BINDER} & - & 74.4 & - & - & - & - \\
& ToG\,+\,GPT-4 \cite{ToG}& - & 82.6 & - & - & 69.5 & - \\
& RoG  \cite{RoG}& 70.8 & 85.7 & 80.0 & 56.2 & 62.6 & 57.8 \\
& PoG  \cite{PoG} & - & 87.3 & - & - & \textbf{75.0} &- \\
& EffiQA  \cite{EffiQA} & - & 82.9 & - & - & 69.5 & - \\
& LightPROF \cite{LightPROF} & - & 83.8 & - & - & 59.3 & - \\
\midrule
\multirow{2}{*}{GNN\,+\,LLM}
& G-Retriever  \cite{G-Retriever}& - & 73.8 & - & - & - & - \\
& GNN-RAG \cite{GNN-RAG}& 71.3 & 85.7 & 80.6 & 59.4 & 66.8 & 61.7 \\
\midrule
\multirow{3}{*}{Ours}
& \modelname (Qwen-max) & \underline{74.7} & \textbf{90.3} & \underline{83.2} & 61.6 & 71.8 & \underline{63.6} \\
& \modelname (GPT-4o-mini) & 69.0 & \underline{89.4} & 80.0 & 53.7 & \underline{72.3} & 62.1 \\
& \modelname (GPT-4-turbo) & \textbf{76.2} & 88.7 & \textbf{83.4} & \textbf{62.3} & 70.8 & \textbf{63.9} \\
& \modelname (GPT-4) & 73.4	& 89.0&	81.4&	\underline{61.9}&	72.2&	63.0 \\
\bottomrule
\end{tabular}}
\end{table*}

\paragraph{Baselines}

We compare \modelname with four categories of methods:
(i) \textbf{Embedding-based.}
KV-Mem \cite{KV-Mem} uses a key–value memory network for KGQA.
EmbedKGQA \cite{EmbedKGQA} leverages pre-trained embeddings for multi-hop reasoning.
TransferNet \cite{TransferNet} improves reasoning within the relation set.
Rigel \cite{Rigel} enhances reasoning for questions involving multiple entities.
(ii) \textbf{GNN-based.}
GraftNet \cite{GraftNet} employs a convolutional GNN.
PullNet \cite{PullNet} builds on GraftNet and learns to retrieve nodes via shortest paths to answers.
NSM \cite{NSM} adapts GNNs for KGQA, and NSM+h \cite{NSM+h} improves multi-hop reasoning.
SQALER \cite{SQALER} selects which facts to retrieve during GNN reasoning.
SR+NSM (+E2E) \cite{SR+NSM} proposes relation-path retrieval.
ReaRev+LMsr \cite{ReaRev} explores diverse reasoning paths in a multi-stage manner.
(iii) \textbf{KG-enhanced LLM-based.}
KD-CoT \cite{KD-CoT} augments chain-of-thought prompting with KG facts.
StructGPT \cite{StructGPT} retrieves KG facts for RAG.
KB-BINDER \cite{KB-BINDER} improves reasoning via logical forms.
ToG \cite{ToG} selects relevant facts step-by-step with a strong LLM.
RoG \cite{RoG} finetunes an LLM to generate relation paths for planning.
PoG \cite{PoG} performs self-correcting planning via sub-goals and reflection.
EffiQA \cite{EffiQA} creates sub-questions and pseudo answers with LLMs to guide a lightweight plug-in retriever over the KG.
LightPROF~\cite{LightPROF} encodes KG structure into soft prompts via a lightweight adapter.
(iv) \textbf{GNN-enhanced LLM-based.}
G-Retriever \cite{G-Retriever} uses GNN-based prompt tuning to assist LLMs.
GNN-RAG \cite{GNN-RAG} integrates a GNN for KG retrieval and finetunes an LLM for question answering.

\paragraph{Evaluation protocol}
We follow previous works \cite{RoG, ToG, GNN-RAG} to adopt F1, Hit, and H@1 as the evaluation metrics.
F1 measures the overall quality of answers by balancing the precision and recall of the predictions.
Hit checks for the presence of at least one correct answer among the final predictions.
H@1 calculates the proportion of instances where one correct answer is the first predicted entity.
We also adopt accuracy as a metric in CSQA, OBQA, MedQA, and ExplaGraphs. 

\subsection{RQ1: Performance}
\label{sec:main_results}

\paragraph{Main results}
Table~\ref{tab:main_results} presents the main experimental results.
We can observe the following findings: 
(i) \modelname outperforms existing methods on most metrics across the four LLMs, particularly when using Qwen-max and GPT-4-turbo as the LLMs. 
This indicates that our approach is versatile across state-of-the-art LLMs. 
(ii) Among the various models, GPT-4-turbo exhibits the strongest performance, followed by Qwen-max and GPT-4, with GPT-4o-mini ranking last. 
Our method is influenced by both the LLM and KGFR, and we anticipate that our method can improve as more advanced models become available in the future. 
Notably, even the smallest GPT-4o-mini model achieves commendable scores. 
(iii) The WebQSP dataset is relatively simple, with the maximum hop of 2, whereas CWQ is more challenging. 
Comparing the results on both datasets, GPT-4o-mini shows a larger gap in F1 scores on CWQ compared to the other two variants. 
This difference arises because more complex questions demand higher levels of language understanding and decision-making capabilities from the LLMs.
(iv) Despite GPT-4o-mini's lower F1 score on CWQ, it achieves a high Hit rate. 
This can be attributed to the more advanced models, Qwen-max and GPT-4-turbo, which tend to be more conservative in their responses. 
In contrast, GPT-4o-mini tends to include more potential candidates.

Overall, these results confirm that LLM–KGFR consistently outperforms prior KGQA systems, scales across diverse LLMs, and remains robust on complex reasoning tasks.

\paragraph{Comparison under unified LLM settings}
To ensure fairness, we align all methods under the same LLM backbones.  
We use two representative models—Llama2-7B and GPT-4—covering most baselines.  
Finetuned or LoRA-based models (e.g., KD-CoT, RoG, G-Retriever, GNN-RAG) use Llama2-7B.  
Frozen-LLM methods include ToG, PoG, EffiQA (GPT-4) and LightPROF (frozen Llama2-7B with a trainable adapter).  
Table~\ref{tab:fair_llm_comparison} reports the results.
For LightPROF, we report its original Llama2-7B results here, while Table~\ref{tab:main_results} reports Llama3-8B.
Without any finetuning, \modelname shows strong performance.  
Under frozen Llama2-7B, it clearly outperforms the frozen and LoRA-finetuned baselines.  
With GPT-4, it reports high F1 and competitive Hit scores—slightly above ToG and EffiQA, and close to PoG.  
These results indicate that our gains mainly come from the retrieval–reasoning synergy rather than LLM finetuning or model size.

\begin{table}[!h]
\setlength{\tabcolsep}{6pt}
\centering
\caption{Comparison under unified LLM settings}
\label{tab:fair_llm_comparison}
{\small
\begin{tabular}{lcccc}
\toprule
\multirow{2}{*}{Method} 
& \multicolumn{2}{c}{WebQSP} 
& \multicolumn{2}{c}{CWQ} \\
\cmidrule(lr){2-3} \cmidrule(lr){4-5}
& F1 & Hit & F1 & Hit \\
\midrule
&\multicolumn{4}{c}{\textit{Llama2-7B (LoRA / Finetuned)}} \\[-2pt]
\midrule
KD-CoT (LoRA) & 52.5 & 68.6 & -- & 55.7 \\
G-Retriever (LoRA) & -- & 73.8 & -- & -- \\
RoG (Finetuned) & -- & 85.7 & 56.2 & 62.6 \\
GNN-RAG (Finetuned) & 71.3 & 85.7 & 59.4 & 66.8 \\
\midrule
&\multicolumn{4}{c}{\textit{Llama2-7B (Frozen)}} \\[-2pt]
\midrule
G-Retriever (w/o LoRA) & -- & 70.5 & -- & -- \\
LightPROF & -- & 71.2 & -- & 48.5 \\
\textbf{\modelname} & 55.7 & 80.8 & 44.6 & 58.1 \\
\midrule
&\multicolumn{4}{c}{\textit{GPT-4 (Frozen)}} \\[-2pt]
\midrule
ToG & -- & 82.6 & -- & 69.5 \\
PoG & -- & 87.3 & -- & 75.0 \\
EffiQA & -- & 82.9 & -- & 69.5 \\
\textbf{\modelname} & 73.4 & 89.0 & 61.9 & 72.2 \\
\bottomrule
\end{tabular}}
\end{table} 

\subsection{RQ2: Generalization Evaluation}
\label{sec:generalization}

We conduct experiments to validate the generalization capabilities of \modelname. 

\begin{figure}[!h]
\centering
  \includegraphics[width=0.6\columnwidth]{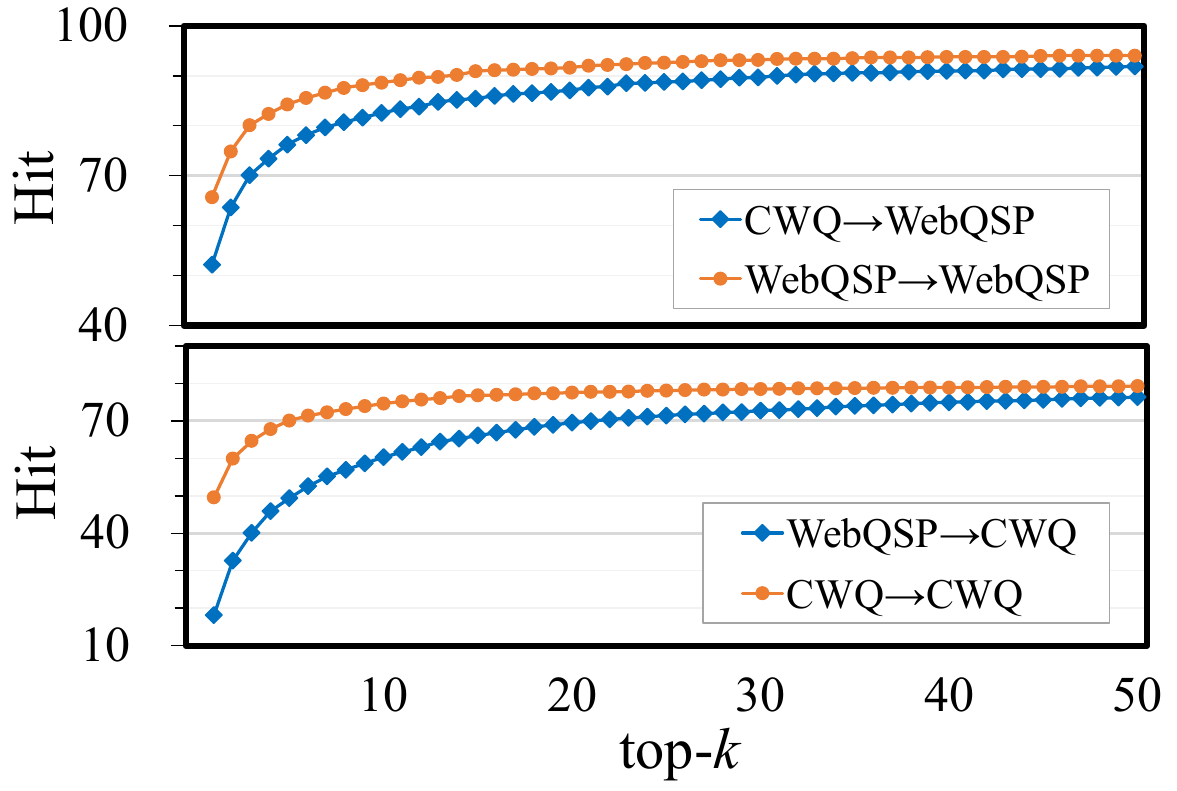}
  \vspace{-4pt}
  \caption{KGFR only top-$k$ retrieval results between WebQSP and CWQ}
  \label{fig:generalization_CWQ_WebQSP}
\end{figure}

\paragraph{Cross-dataset transfer between WebQSP and CWQ}
To explore the generalization and knowledge transfer capabilities of \modelname, 
we pre-train two KGFRs using WebQSP and CWQ, respectively. 
We then test the two models on both datasets, resulting in four knowledge transfer settings. 
Figure~\ref{fig:generalization_CWQ_WebQSP} shows the Hit scores w.r.t. various values of $k$.

We observe a slight performance gap when comparing results from training and testing on the same dataset versus different datasets, and this gap continues to narrow as $k$ increases.
Notably, the gap is smaller in the ``CWQ\,$\rightarrow$\,WebQSP'' case compared to ``WebQSP\,$\rightarrow$\,CWQ''. 
This discrepancy can be attributed to the significantly larger training set for CWQ, which consists of 27,639 samples, compared to only 2,848 samples in WebQSP.
Furthermore, we conduct experiments on the complete \modelname based on these combinations, and the results are shown in Table~\ref{tab:result_generalization_WebQSP_CWQ}. 
This gap further narrows after collaborating with an LLM.

\begin{table}[!ht]
\setlength{\tabcolsep}{4pt}
\centering
\caption{Generalization between WebQSP and CWQ}
\label{tab:result_generalization_WebQSP_CWQ}
{\small
\begin{tabular}{ll|ccc}
\toprule
LLMs & Training\,$\rightarrow$\,Testing & F1 & Hit & H@1 \\
\midrule
\multirow{4}{*}{Qwen-max}
& CWQ\,$\rightarrow$\,WebQSP & 72.9 & 86.4 & 81.8 \\
& WebQSP\,$\rightarrow$\,WebQSP & 74.7 & 90.3 & 83.2 \\
\cmidrule(lr){2-5}
& WebQSP\,$\rightarrow$\,CWQ & 57.6 & 70.2 & 62.1 \\
& CWQ\,$\rightarrow$\,CWQ & 61.6 & 71.8 & 63.6 \\
\midrule
\multirow{4}{*}{GPT-4o-mini}
& CWQ\,$\rightarrow$\,WebQSP & 68.9 & 88.6 & 79.2 \\
& WebQSP\,$\rightarrow$\,WebQSP & 69.0 & 89.4 & 80.0 \\
\cmidrule(lr){2-5}
& WebQSP\,$\rightarrow$\,CWQ & 52.3 & 70.5 & 61.2 \\
& CWQ\,$\rightarrow$\,CWQ & 53.7 & 72.3 & 62.1 \\
\bottomrule
\end{tabular}}
\end{table} 

\paragraph{Generalization to unseen dataset}

We further validate \modelname's generalization on GrailQA~\cite{GrailQA} (an entirely unseen dataset during pre-training).
The evaluation setup and the LLM are the same as PoG~\cite{PoG}.
The results are shown in Table~\ref{tab:GrailQA}.
\modelname maintains strong knowledge transfer and achieves competitive performance on GrailQA, demonstrating its robust generalization capabilities. 
These results further confirm that \modelname can effectively handle novel questions in unseen domains.

\begin{table}[!h]
\centering
\caption{Performance on the GrailQA dataset}
\label{tab:GrailQA}
\begin{tabular}{l|cccc}
\toprule
Methods & Overall & I.I.D & Compositional & Zero-shot \\ \midrule
ToG & 68.7 & 70.1 & 56.1 & 72.7 \\
PoG & 76.5 & 76.3 & 62.1 & 81.7 \\
\modelname & \textbf{80.2} & \textbf{79.4} & \textbf{74.8} & \textbf{82.4} \\ \bottomrule
\end{tabular}
\end{table} 

\begin{figure}[!h]
    \vskip -5pt
    \centering
    \includegraphics[width=0.7\linewidth]{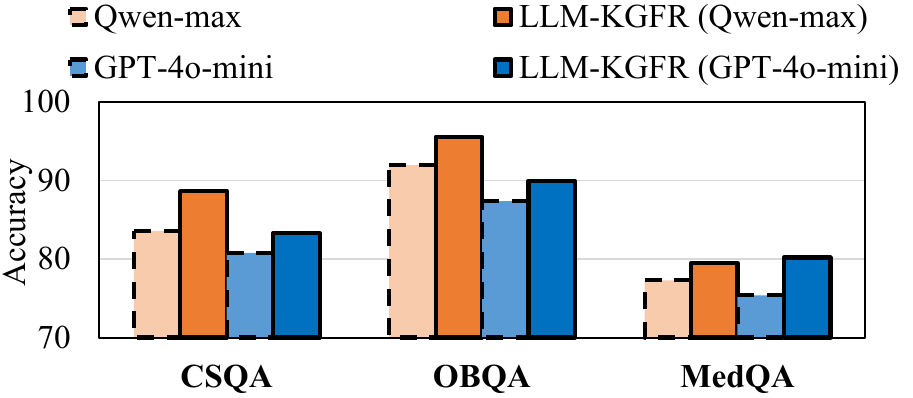}
    \caption{Accuracy of multiple-choice QA}
    \label{fig:generalization_multi_choice}
\end{figure}

\paragraph{Generalization to multiple-choice QA}
We evaluate the multiple-choice QA performance using the CSQA, OBQA, and MedQA datasets. 
The background KGs for CSQA and OBQA are sourced from ConceptNet, while MedQA uses UMLS and DrugBank. 
We contrast a direct LLM-answering baseline with \modelname.
For \modelname, we incorporate the options as known information and ask the LLM to return the most likely answer option.
The results are displayed in Figure~\ref{fig:generalization_multi_choice}. 
\modelname achieves consistent improvements across all LLMs and datasets compared to direct answering.

We further evaluate \modelname against state-of-the-art GNN-based approaches to establish comprehensive benchmarks. 
Table~\ref{tab:gnn_comparison} compares the performance on three challenging QA datasets (CSQA, OBQA, and MedQA) using the results from the PaperWithCode leaderboard\footnote{\url{https://paperswithcode.com/}}.
The results reveal that \modelname achieves superior performance across all benchmarks, particularly showing remarkable gains 
on the biomedical MedQA dataset. 
This advantage stems from three key factors: 
(i) The KGFR's robust knowledge retrieval capabilities that surpass traditional graph propagation methods. 
(ii) The LLM's strong language understanding and reasoning capabilities. 
(iii) Their synergistic interaction that dynamically adapts to different question types. 
The consistent superiority across both general-domain (CSQA/OBQA) and specialized (MedQA) benchmarks demonstrates \modelname's versatility compared to previous GNN-based approaches.

\begin{table}[!ht]
\centering
\caption{Accuracy comparison with GNN-based methods across multiple QA datasets}
\label{tab:gnn_comparison}
\begin{tabular}{l|ccc}
\toprule
Methods & CSQA & OBQA & MedQA \\ 
\midrule
QA-GNN~\cite{QA-GNN} & 76.5 & 82.8 & 38.0 \\
DEKCOR~\cite{DEKCOR} & 83.3 & 82.4 & -- \\
DRAGON~\cite{DRAGON} & 72.0 & 76.0 & -- \\
GSC~\cite{GSC} & 79.1 & 87.4 & -- \\
GrapeQA~\cite{GrapeQA} & 74.9 & 90.0 & 39.5 \\
GNR~\cite{GNR} & -- & 89.6 & -- \\ \midrule
\modelname (GPT-4o-mini) & 83.3 & 90.0 & \textbf{80.2} \\
\modelname (Qwen-max) & \textbf{88.6} & \textbf{95.6} & 79.5 \\ 
\bottomrule
\end{tabular}
\end{table}

\paragraph{Generalization to commonsense reasoning.}
ExplaGraphs is a dataset for generative commonsense reasoning, evaluating whether arguments are supportive or contradictory.
Its background graphs are small enough to fit entirely within a single prompt, making it ideal for testing structured retrieval effects beyond context-length limits.
We evaluate four variants: (i) a pure LLM incorporating the background KG into prompts, (ii) our full \modelname that retrieves relevant entities and verbalizes corresponding facts into natural sentences, (iii) \modelname (w/o retrieval), and (iv) \modelname (w/o verbalization).
Although the answers are not entities, we still retrieve and rank relevant entities, and the relation verbalization module converts retrieved facts into coherent natural-language evidence.
Results are shown in Table~\ref{tab:result_explagraph}.
Qwen-max already surpasses existing baselines, while our structured retrieval and fact verbalization further boost performance.
Even though the graphs can be fully included in prompts, \modelname still shows consistent gains, indicating that improvements stem from structured retrieval and semantic verbalization rather than context truncation.
Ablation shows that removing verbalization drops accuracy from 94.3 to 93.9, and removing retrieval further to 93.3, confirming that both components are complementary and essential.

\begin{table}[!h]
\centering
\caption{Accuracy on ExplaGraphs}
\label{tab:result_explagraph}{\small
\resizebox{0.8\linewidth}{!}{
\begin{tabular}{l|c}
\toprule
Methods & Acc. \\
\midrule
Zero-shot \cite{G-Retriever} & 56.5 \\
Zero-CoT \cite{COT} & 57.0 \\
CoT-BAG \cite{CoT-BAG} & 57.9 \\
KAPING \cite{KAPING} & 62.3 \\
GraphToken \cite{GraphToken} & 85.1 \\
G-Retriever \cite{G-Retriever} & 85.2 \\
G-Retriever (LoRA) \cite{G-Retriever} & 87.1 \\
\midrule
Qwen-max & 92.1 \\
\modelname (Qwen-max) & \textbf{94.3} \\
\modelname (Qwen-max w/o verbalization) & 93.9 \\
\modelname (Qwen-max w/o retrieval) & 93.3 \\
\bottomrule
\end{tabular}}
}
\end{table}

\subsection{RQ3: Scalability and Efficiency}
\label{sec:ablation_study}

\paragraph{Scalability analysis}
\label{sec:scalability}

We propose two strategies to ensure scalability: progressive expansion (PE) and asymmetric pruning (AP) in Section~\ref{sec:APP}. 
Here, we conduct experiments to analyze their effects on the message propagation range and to examine the trade-off between efficiency and accuracy.
Specifically, we design five variants based on whether PE and AP are enabled, as well as the threshold $\lambda$.
For each variant, we calculate the average number of entities and facts (facts are treated as directed edges) involved in propagation for each question.
We then evaluate their H@1 scores under the condition without LLM collaboration (w/o LLM).
The experimental results are shown in Table~\ref{tab:pe_ap_threshold}.
``OOM'' denotes Out-of-Memory.

\begin{table}[!ht]
\centering
\caption{Effects of Progressive Propagation (PE) and Asymmetric Pruning (AP) under different thresholds $\lambda$}
\label{tab:pe_ap_threshold}
\resizebox{\linewidth}{!}{
\begin{tabular}{l|cc|c|cc|c}
\toprule
Dataset & PE & AP & $\lambda$ & Entity & Fact & H@1 (w/o LLM) \\
\midrule
\multirow{5}{*}{WebQSP} 
 & $\times$ & $\times$ & - & \ 1.3m & \ 7.6m & OOM \\
 & $\checkmark$ & $\times$ & - & \ 0.6m & \ 3.4m & OOM \\
 & $\checkmark$ & $\checkmark$ & 1,000 & \ 0.2m  & \ 1.5m  & 65.4 \\
 & $\checkmark$ & $\checkmark$ & 100 & \ 0.1m & \ 0.6m & 65.7 \\
 & $\checkmark$ & $\checkmark$ & 10 & \ 9.8k & 44.8k & 60.2 \\
\midrule
\multirow{5}{*}{CWQ} 
 & $\times$ & $\times$ & - & \ 2.3m & 14.5m & OOM \\
 & $\checkmark$ & $\times$ & - & \ 0.8m & \ 5.1m & OOM \\
 & $\checkmark$ & $\checkmark$ & 1,000 & \ 0.3m &  \ 1.6m & 49.8 \\
 & $\checkmark$ & $\checkmark$ & 100 & \ 0.1m & \ 0.6m & 49.7 \\
 & $\checkmark$ & $\checkmark$ & 10 & \ 7.9k & 38.8k & 38.4 \\
\bottomrule
\end{tabular}
}
\end{table}

Without PE or AP, propagation expands to millions of nodes and edges, leading to OOM errors.
PE alone reduces the propagation range by progressively expanding neighborhoods, but still suffers from scalability issues.
When both PE and AP are enabled, the propagation size is significantly reduced while maintaining comparable accuracy.
As shown in Table~\ref{tab:pe_ap_threshold}, decreasing $\lambda$ from 1000 to 100 substantially reduces the number of involved entities and facts, while the H@1 score remains nearly unchanged on both datasets.
However, when $\lambda$ is further reduced to 10, the propagation range becomes much smaller, accompanied by a noticeable drop in H@1.
These results indicate that a moderate threshold (e.g., $\lambda{=}100$) provides a favorable balance between efficiency and accuracy.

\begin{figure}[!ht]
    \centering
    \begin{minipage}[b]{0.441\linewidth}
        \centering
        \includegraphics[width=\linewidth]{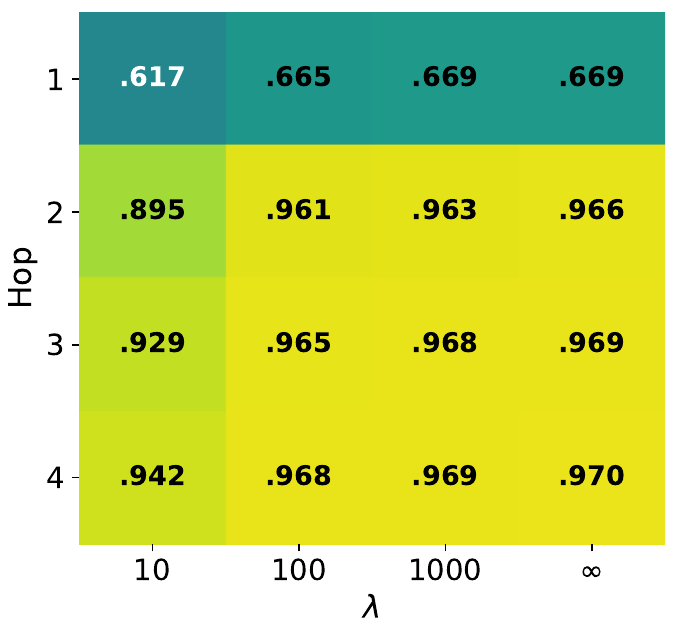}
        {$\ \ \ \ \ $\small (a) WebQSP}
    \end{minipage}
    \hfill
    \begin{minipage}[b]{0.539\linewidth}
        \centering
        \includegraphics[width=\linewidth]{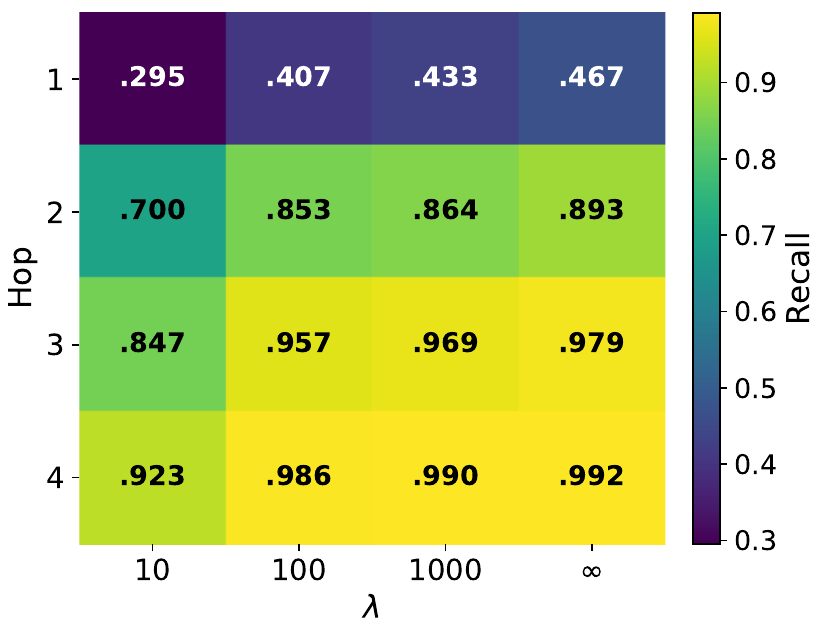}
        {\small (b) CWQ$\ \ \ $}
    \end{minipage}
    
    \caption{Answer reachability under different thresholds $\lambda$ and hop limits. 
    Each cell shows the proportion of questions for which the ground-truth answer remains reachable under constrained expansion.}
    \label{fig:lambda_heatmap}
\end{figure}

To further examine whether pruning affects answer coverage, we analyze answer reachability under different $\lambda$ values and hop limits.
For each question, we check whether the ground-truth answer is reachable from at least one topic entity within the constrained propagation subgraph.
The results are shown in Figure~\ref{fig:lambda_heatmap}.
We observe that reachability increases with larger $\lambda$ but quickly saturates, and $\lambda{=}100$ already achieves coverage close to the less constrained setting.
In contrast, overly aggressive pruning begins to exclude necessary reasoning paths, leading to reduced coverage.
This suggests that the impact of pruning on recall is limited within a reasonable range of $\lambda$, and further supports the use of a moderate threshold.

\paragraph{Efficiency analysis}
\label{sec:efficiency}

We conduct experiments to analyze the inference efficiency of \modelname. 
We compare the LLM calls, token usage and inference time with state-of-the-art black-box models, while excluding LLM-tuning-based approaches that require costly LLM finetuning for each dataset.
Our message-passing component completes most retrievals within 0.5 second, while our LLM generation phase typically requires only 2--6 total operations per question (combining generation and reflection steps).
As shown in Table~\ref{tab:efficiency}, \modelname substantially outperforms existing methods across all efficiency metrics.
These efficiency improvements are most evident when handling complex queries, where our KGFR retrieval mechanism eliminates the need for LLM calls required by LLM-based retrieval.

\begin{table}[h]
\centering
\caption{Efficiency comparison with baseline methods}
\label{tab:efficiency}
\begin{tabular}{l|l|ccc}
\toprule
Datasets & Methods & LLM calls & Tokens & Time (s) \\
\midrule
\multirow{3}{*}{CWQ} 
    & ToG & 22.6 & 9,669.4 & 96.5 \\
    & PoG & 13.3 & 8,156.2 & 23.3 \\
    & \modelname & \ \,2.4 & 3,145.1 & \ \,9.2 \\
\midrule
\multirow{3}{*}{WebQSP} 
    & ToG & 15.9 & 7,018.9 & 63.1 \\
    & PoG & \ \,9.0 & 5,517.7 & 16.8 \\
    & \modelname & \ \,2.1 & 2,725.2 & \ \,7.6 \\
\bottomrule
\end{tabular}
\end{table}

\subsection{RQ4: Further Analysis}
\label{sec:RQ4}

\paragraph{Ablation Study}
We conduct an ablation study to evaluate the contribution of each module.  
Specifically, six variants of \modelname (Qwen-max) are constructed by selectively removing key components:  
``w/o LLM ($k$=10)’’ removes the LLM and directly ranks the top-10 entities;  
``w/o description’’ replaces generated relation descriptions with raw relation names;  
``w/o node-/edge-/path-level retrieval’’ respectively disable different hierarchical retrieval interfaces;  
and ``w/o reflection’’ removes the reflection loop.
Table~\ref{tab:ablation_study} shows that the complete \modelname achieves the highest F1 and H@1 scores across both datasets, demonstrating that all modules positively contribute to overall performance.  
Removing the LLM causes a dramatic F1 drop, confirming that language-guided reasoning and answer synthesis are indispensable.  
The ``w/o description’’ variant performs close to the full model since raw relation names still carry partial semantic cues from pre-training.
Among retrieval interfaces, the node-level retrieval yields the largest performance degradation when removed, validating it as the core of KGFR’s reasoning process.  
Edge- and path-level retrievals also bring steady gains by enriching multi-hop context and refining fact-level evidence.    
Finally, removing the reflection module slightly reduces accuracy, since the pre-retrieval stage already offers strong candidate and fact grounding, though reflection further enhances stability and prevents premature termination in complex cases.

\begin{table}[!ht]
\centering
\caption{Ablation results}
\label{tab:ablation_study}
\resizebox{\linewidth}{!}{
\begin{tabular}{l|ccc|ccc}
\toprule
\multirow{2}{*}{Methods}
& \multicolumn{3}{c|}{WebQSP} & \multicolumn{3}{c}{CWQ} \\ 
\cmidrule(lr){2-4} \cmidrule(lr){5-7}
& F1 & Hit & H@1 & F1 & Hit & H@1 \\ 
\midrule
\modelname (Qwen-max) & 74.7 & 90.3 & 83.2 & 61.6 &	71.8 &	63.6 \\
\quad w/o LLM ($k=10$) & 29.5 & 88.5 & 65.7 & 18.0 & 74.7 & 49.7 \\
\quad w/o description & 71.6 & 85.9 & 81.2 & 61.2 & 69.6 & 62.9 \\
\quad w/o node-level retrieval & 62.7 & 73.8 & 68.6 & 51.3 & 56.4 & 54.0\\
\quad w/o edge-level retrieval & 69.3 & 87.3 & 81.4 & 58.9 & 70.1 & 62.4 \\
\quad w/o path-level retrieval & 66.2 & 86.1 & 80.2 & 57.3 & 69.8 & 62.2\\
\quad w/o reflection & 71.9 & 89.5 & 81.4 & 59.7 & 69.9 & 62.0 \\
\bottomrule
\end{tabular}
}
\end{table}

\paragraph{Effect of encoder choice}
\label{sec:encoder_effect}

To further analyze the influence of the textual encoder used in relation and question representation, 
we fix the LLM to Qwen-max and replace the default BGE-Large-EN-v1.5 with several alternative BERT-based encoders of different sizes and training objectives, 
including BERT-base/large, SentenceBERT-base/large, and BGE-base/large.
Table~\ref{tab:encoder_comparison} reports the results on WebQSP and CWQ.

\begin{table}[!ht]
\centering
\caption{Performance comparison of \modelname\ with different BERT encoders.}
\label{tab:encoder_comparison}
\resizebox{\linewidth}{!}{
\begin{tabular}{l|ccc|ccc}
\toprule
\multirow{2}{*}{Encoder} & \multicolumn{3}{c|}{WebQSP} & \multicolumn{3}{c}{CWQ} \\
\cmidrule(lr){2-4} \cmidrule(lr){5-7}
 & F1 & Hit & H@1 & F1 & Hit & H@1 \\
\midrule
BERT-base & 70.8 & 84.0 & 80.2 & 58.5 & 67.9 & 59.6 \\
BERT-large & 72.6 & 89.7 & 82.6 & 60.7 & 71.0 & 62.2 \\
SentenceBERT-base & 71.4 & 85.3 & 80.3 & 59.4 & 68.1 & 60.3 \\
SentenceBERT-large & 74.1 & 88.4 & 82.4 & 61.2 & 71.4 & 62.4 \\
BGE-base & 70.4 & 85.6 & 79.9 & 57.6 & 68.9 & 60.0 \\
BGE-large & \textbf{74.7} & \textbf{90.3} & \textbf{83.2} & \textbf{61.6} & \textbf{71.8} & \textbf{63.6} \\
\bottomrule
\end{tabular}}
\end{table}

We observe that encoder capacity plays a key role in retrieval quality.
Larger encoders (e.g., BERT-large and BGE-large) consistently outperform their base counterparts across both datasets, 
indicating that a stronger language encoder yields more informative question and relation representations.
Sentence-level contrastive training (as in SentenceBERT and BGE) also provides moderate gains over vanilla BERT, 
suggesting that semantic alignment between question and relation text further enhances KGFR's reasoning accuracy.
Overall, these results confirm that the encoder’s representation quality directly impacts the generalization ability and overall performance of the retriever.

\paragraph{Effect of LLM scale and architecture}
\label{app:llm_scale}

In the main experiments, we primarily adopt commercial black-box LLMs (e.g., GPT-4, GPT-4-turbo, Qwen-max).
To further investigate how the scale and architecture of the LLM affect overall performance, 
we replace these models with open-source LLMs of different sizes, including Llama3-8B, Llama3-70B, Qwen2.5-7B, and Qwen2.5-72B.
Table~\ref{tab:open_source_LLMs_results} reports the results on WebQSP and CWQ.
The results show a consistent trend:  
larger LLMs yield stronger accuracy and answer consistency, while smaller ones remain competitive.  
Specifically, the 70B and 72B models achieve improvements of about 6–8 points in F1 over their 7B and 8B counterparts, demonstrating that \modelname can effectively exploit richer linguistic representations from larger models.  
Meanwhile, the solid performance of the smaller models confirms that our retriever–reasoner collaboration remains effective even with lightweight LLMs, 
highlighting \modelname's scalability and robustness across diverse LLM backbones.

\begin{table}[!ht]
\centering
\caption{Performance comparison of \modelname\ with open-source LLMs of different scales.}
\label{tab:open_source_LLMs_results}
\resizebox{\linewidth}{!}{
\begin{tabular}{l|ccc|ccc}
\toprule
\multirow{2}{*}{LLM Backbone} & \multicolumn{3}{c|}{WebQSP} & \multicolumn{3}{c}{CWQ} \\
\cmidrule(lr){2-4} \cmidrule(lr){5-7}
 & F1 & Hit & H@1 & F1 & Hit & H@1 \\
\midrule
\modelname (Llama3-8B) & 66.8 & 82.4 & 75.7 & 47.7 & 61.9 & 52.5 \\
\modelname (Llama3-70B) & 73.1 & 90.1 & 82.0 & 52.5 & 67.8 & 58.1 \\
\modelname (Qwen2.5-7B) & 68.4 & 83.3 & 76.5 & 43.6 & 57.2 & 48.8 \\
\modelname (Qwen2.5-72B) & \textbf{72.9} & \textbf{89.8} & \textbf{82.1} & \textbf{55.4} & \textbf{66.1} & \textbf{57.7} \\
\bottomrule
\end{tabular}}
\end{table}

\paragraph{Impact of relation description generator}
\label{sec:llm_analysis}
To evaluate the impact of relation description generator, 
we replace the description generator with different LLMs.
Specifically, we use Llama3-8B and Llama3-70B, 
while the default setting uses GPT-4o-mini.
The results are shown in Table~\ref{tab:llm_compare}.
We observe that Llama3-70B achieves performance comparable to GPT-4o-mini when used as the description generator. While the performance of Llama3-8B shows a noticeable decline, it remains at a reasonably strong level.
These results suggest that stronger LLMs generally yield better performance, and models at the level of Llama3-70B are already capable of producing sufficiently high-quality descriptions. Moreover, smaller models such as Llama3-8B can still maintain competitive results, primarily because each generated description explicitly includes the original relation name, which provides a reliable lower bound on information quality.

\begin{table}[h]
\centering
\caption{Performance comparison using different LLMs as relation description generators.}
\small
\resizebox{\linewidth}{!}{
\begin{tabular}{l|ccc|ccc}
\toprule
\multirow{2}{*}{Descrip. Generator} & \multicolumn{3}{c|}{WebQSP} & \multicolumn{3}{c}{CWQ} \\
\cmidrule(lr){2-4} \cmidrule(lr){5-7}
 & F1 & Hit@1 & Hit@10 & F1 & Hit@1 & Hit@10 \\
\midrule
Llama3-8B  & 72.1 & 86.4 & 82.0 & 61.0 & 69.8 & 62.3 \\
Llama3-70B & \textbf{74.9} & 89.9 & \textbf{83.6} & 61.3 & 71.0 & 62.9 \\
GPT-4o-mini & 74.7 & \textbf{90.3} & 83.2 & \textbf{61.6} & \textbf{71.8} & \textbf{63.6} \\
\bottomrule
\end{tabular}
}
\label{tab:llm_compare}
\end{table}

\paragraph{Incorporating LLM-based retrieval augmentation}
\label{sec:incorporate_RoG}

We introduce KGFR-based retrieval methods in Section~\ref{sec:collaborative_question_answering}.
They can collaborate with LLM-based retrievers to further enhance reasoning. 
Here, we follow \cite{GNN-RAG} to create some variants of \modelname that incorporate finetunable LLM-based retrieval \cite{RoG}. 
Specifically, during the pre-retrieval, we integrate the results from the LLM-based retrieval into the prompt to strengthen reasoning, with the results shown in Table~\ref{tab:incorporation_with_RoG}.
We observe further performance improvements, which indicate that the information from the KGFR and LLM retrieval is complementary. 
It also demonstrates that \modelname can integrate LLM-based retrieval for further enhancement.

\begin{table}[!h]
\centering
\caption{Results of \modelname with LLM-based retrieval augmentation (RA)}
\vskip 6pt
\resizebox{\linewidth}{!}{
\begin{tabular}{l|ccc|ccc}
\toprule
\multirow{2}{*}{Methods} 
& \multicolumn{3}{c|}{WebQSP} 
& \multicolumn{3}{c}{CWQ} \\ 
\cmidrule(lr){2-4} \cmidrule(lr){5-7}
& F1 & Hit & H@1 & F1 & Hit & H@1 \\
\midrule
GNN-RAG & 71.3 & 85.7 & 80.6 & 59.4 & 66.8 & 61.7 \\
GNN-RAG\,+\,RA & 73.5	& 90.7	& 82.8	& 60.4	&68.7	&62.8\\
\midrule
\modelname (Qwen-max) &74.7 & 90.3 & 83.2 & 61.6 & 71.8 & 63.6 \\
\modelname (Qwen-max)\,+\,RA & 76.3&	89.9&	83.8&	62.9&	72.7&	64.2 \\
\midrule
\modelname (GPT-4o-mini)& 69.0 & 89.4 & 80.0 & 53.7	& 72.3 & 62.1 \\
\modelname (GPT-4o-mini)\,+\,RA & 72.4 &	92.0 &83.1&	54.6& 72.2&	62.4\\
\bottomrule
\end{tabular}
}
\label{tab:incorporation_with_RoG}
\end{table}

\subsection{Case Study}
\label{sec:case_study}

\begin{table*}
\centering
\caption{Qualitative case studies illustrating success and failure modes of the proposed framework.}
\small
\begin{tabular}{c|p{4cm}|p{10.5cm}}
\toprule
\textbf{Step} & \textbf{Stage} & \textbf{Operation / Output} \\
\midrule
\multicolumn{3}{c}{\textbf{(a) Failure Case}} \\
\midrule
0 & Question and Answer & Q: What person notable with ADHD is related to Scarlett Johansson? \\
  &       & A: Justin Timberlake (unseen) \\
1 & Candidate Retrieval & Top candidates: Justin Timberlake, Ryan Reynolds, Hilary Duff, $\dots$ \\
2 & Initial Inference & Output: Justin Timberlake, Bill Cosby, ... \\
3 & Reflection Trigger & Subquestion: “What notable people have ADHD?” \\
4 & Subquestion Answering & Retrieved: Bill Cosby, $\dots$ \\
5 & Second Inference & Output: Bill Cosby (Justin Timberlake removed) \\
6 & Final Answer & Bill Cosby \\
\midrule
\multicolumn{3}{c}{\textbf{(b) Success Case}} \\
\midrule
0 & Question and Answer & Q: What is the capital city of the nation using the South Korean won? \\
  &       & A: Seoul (unseen) \\
1 & Candidate Retrieval & Top candidates: South Korea, Asia, Constitutional republic, ... \\
2 & Initial Inference & Output: South Korea \\
3 & Reflection Trigger & Subquestion: “What is the capital city of South Korea?” \\
4 & Subquestion Answering & Retrieved: Seoul, ... \\
5 & Second Inference & Output: Seoul \\
6 & Final Answer & Seoul \\
\bottomrule
\end{tabular}
\label{tab:case_study}
\end{table*}

To better understand the behavior of the proposed framework, we present qualitative case studies in Table~\ref{tab:case_study}, including both a success case and a failure case.
In the failure example, although the correct answer appears in the initial candidate set, the reflection step generates a subquestion that does not preserve the original query constraints. This leads to a shift toward a broader but less relevant candidate space, causing the model to discard the correct answer and produce an incorrect prediction.
In contrast, the success example shows that when the generated subquestion remains aligned with the original query, the retrieval process can focus on relevant evidence and successfully recover the correct answer.
These results indicate that preserving query constraints during subquestion generation is critical for robust multi-step reasoning, suggesting a direction for further improvement.

\section{Conclusions}
\label{sec:conclusions}

We present \modelname, a collaborative framework that unifies LLMs with a KG retriever for generalized and scalable KGQA.
Through language-guided initialization and asymmetric progressive propagation, KGFR achieves efficient retrieval and strong cross-KG generalization without finetuning.
With multi-level retrieval and reflection-based reasoning, the framework enables controllable, interpretable question answering.
Extensive experiments on seven benchmarks verify that LLM–KGFR consistently outperforms existing methods in accuracy, efficiency, and transferability.
Future work will explore semantic-aware pruning strategies to better balance efficiency and recall, and extend this paradigm to broader unstructured knowledge sources for more comprehensive reasoning.

\bibliographystyle{IEEEtran}
\bibliography{custom}


\begin{IEEEbiography}[{\includegraphics[width=1in,height=1.25in,clip,keepaspectratio]{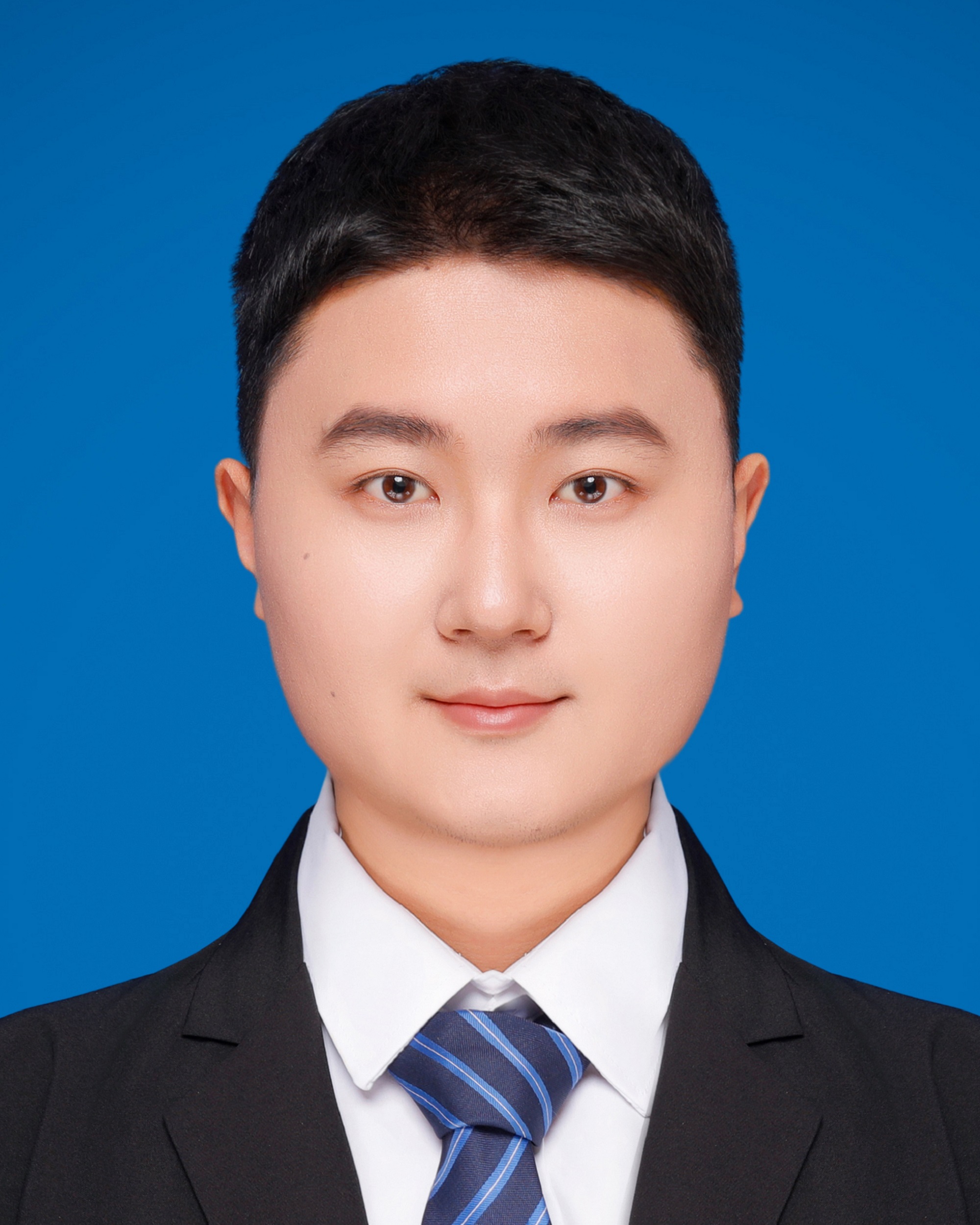}}]{Yuanning Cui} received the BS and MS degrees in Computer Science and Technology from China University of Mining and Technology in 2018 and Nanjing University of Aeronautics and Astronautics in 2021, respectively, and the PhD degree from Nanjing University in 2025. He is currently a Lecturer with the School of Computer, Nanjing University of Information Science and Technology, China. His research interests include knowledge graphs, foundation models, and question answering.
\end{IEEEbiography}

\begin{IEEEbiography}[{\includegraphics[width=1in,height=1.25in,clip,keepaspectratio]{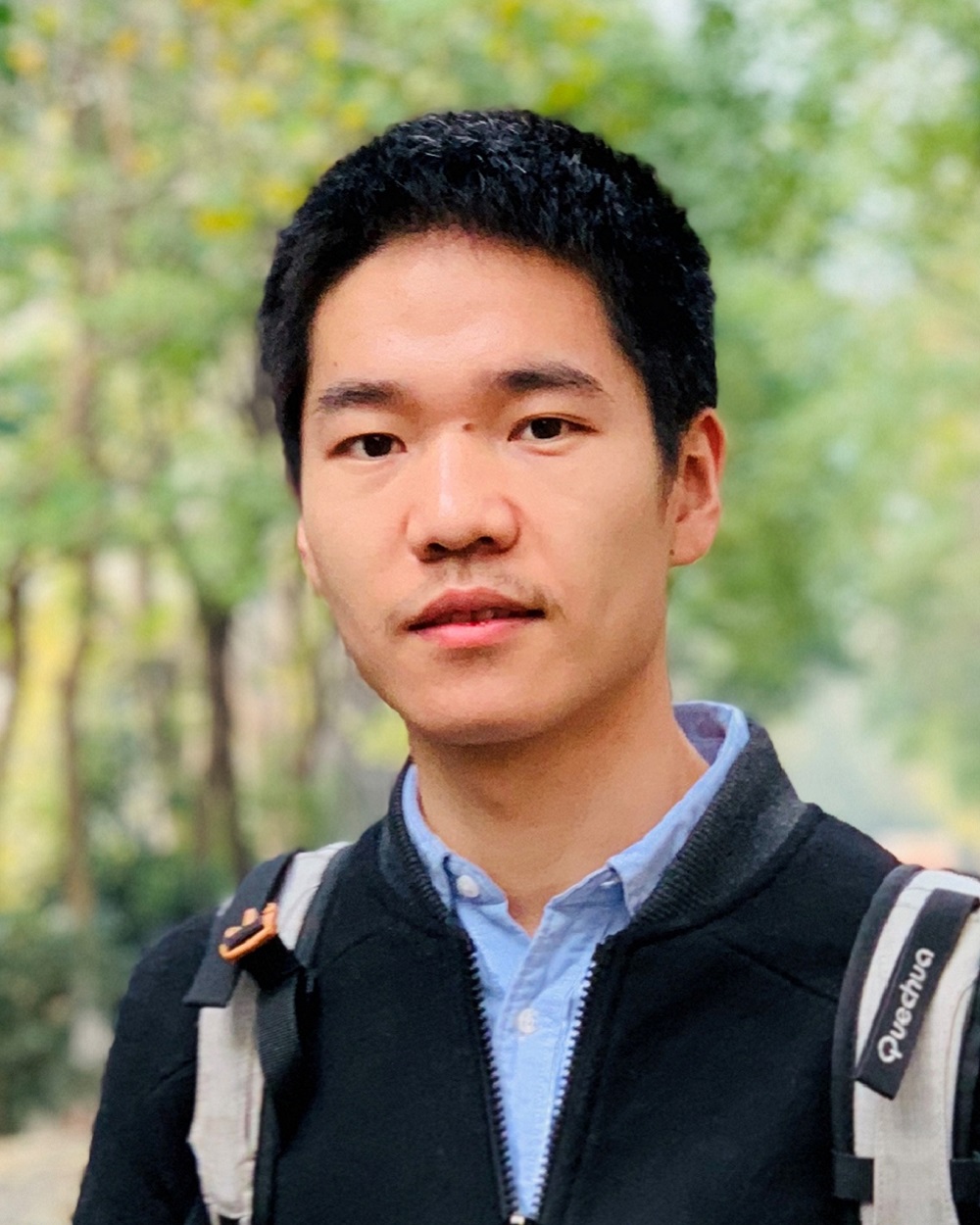}}]{Zequn Sun} is currently an assistant professor at Nanjing University, China. He received his BS degree in Computer Science and Technology from Hohai University, China, in 2016, and PhD degree in Computer Science and Technology from Nanjing University, China, in 2023. His research interests include knowledge graph, representation learning, and entity alignment.
\end{IEEEbiography}

\begin{IEEEbiography}[{\includegraphics[width=1in,height=1.25in,clip,keepaspectratio]{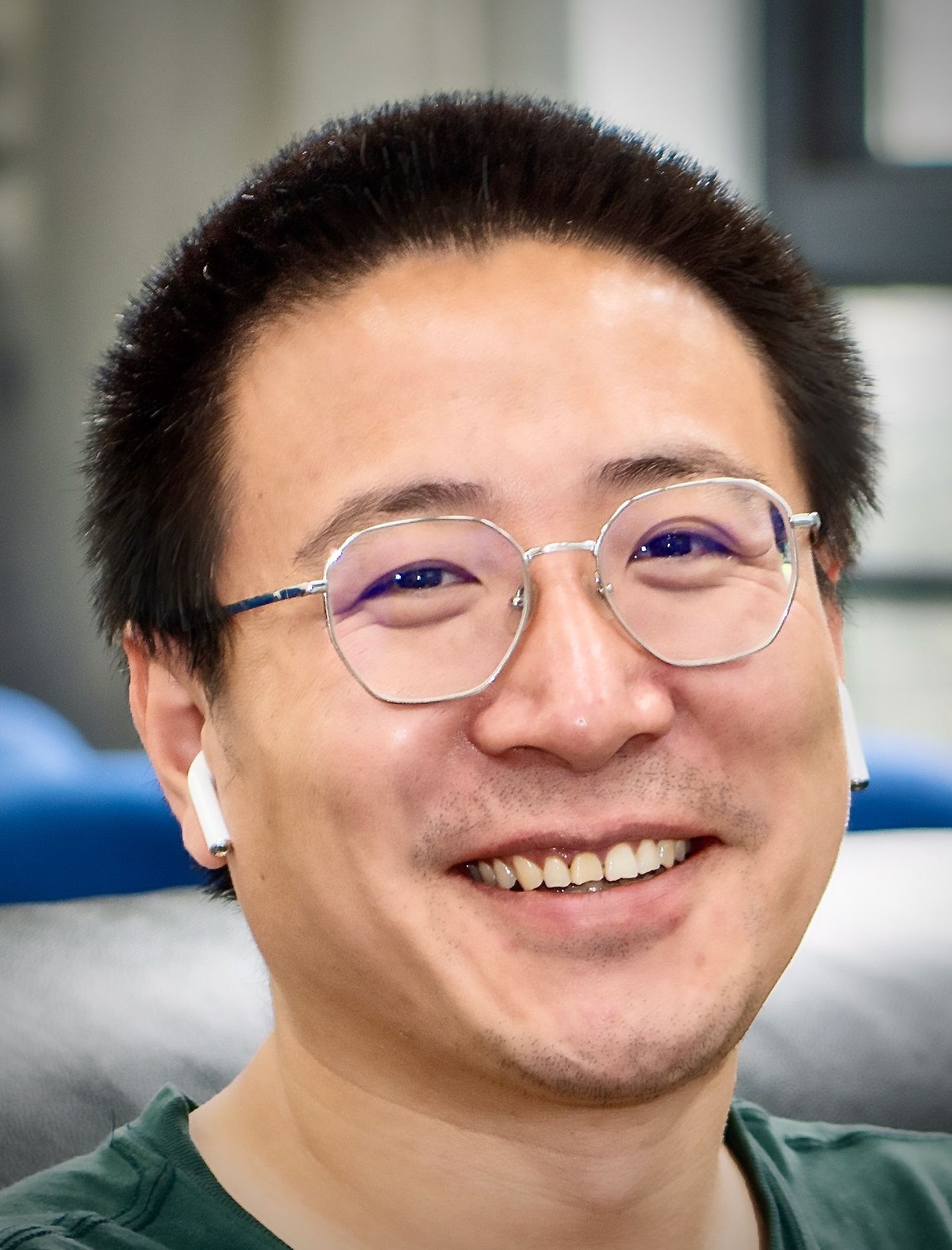}}]{Wei Hu} is a full professor at the State Key Laboratory for Novel Software Technology and the National Institute of Healthcare Data Science, Nanjing University, China. 
He received his PhD degree in Computer Software and Theory in 2009, and BS degree in Computer Science and Technology in 2005, both from Southeast University, China. 
His main research interests include knowledge graph, database, and intelligent software.
\end{IEEEbiography}

\begin{IEEEbiography}[{\includegraphics[width=1in,height=1.25in,clip,keepaspectratio]{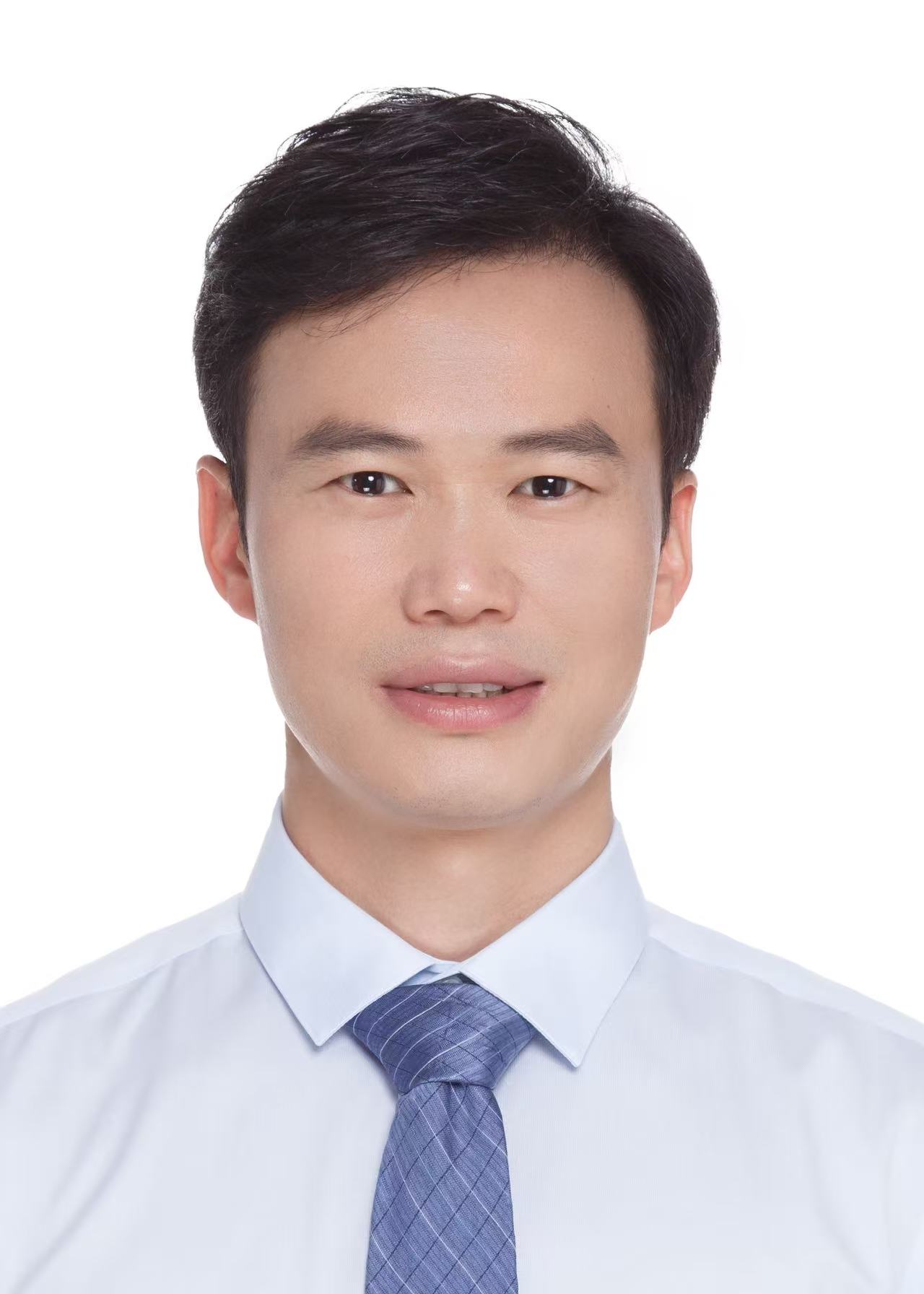}}]{Zhangjie Fu} (Member, IEEE) received the Ph.D. degree in computer science from the College of Computer, Hunan University, China, in 2012. He is currently a Professor with the School of Cyber Science and Engineering, Nanjing University of Information Science and Technology, China. His research interests include cloud and outsourcing security, digital forensics, networks, and information security. His research has been supported by NSFC, PAPD, and GYHY. He is a member of ACM.
\end{IEEEbiography}

\vfill

\end{document}